\documentclass[a4paper,fleqn]{cas-sc}

\usepackage[authoryear,longnamesfirst]{natbib}

\usepackage{subcaption} 
\usepackage{algorithm}
\usepackage{algcompatible}
\usepackage{bm} 
\usepackage{url} 
\usepackage{ltablex} 

\def\tsc#1{\csdef{#1}{\textsc{\lowercase{#1}}\xspace}}
\tsc{WGM}
\tsc{QE}
\tsc{EP}
\tsc{PMS}
\tsc{BEC}
\tsc{DE}

\begin{document}
\let\WriteBookmarks\relax
\def\floatpagepagefraction{1}
\def\textpagefraction{.001}
\shorttitle{A Novel Deep Implicit Imitation RL Framework}
\shortauthors{I. Chrysomallis \& G. Chalkiadakis}

\title [mode = title]{
Going Beyond Expert Performance \\ via Deep Implicit Imitation Reinforcement Learning 
}
\tnotemark[1]

\tnotetext[1]{The research described in this paper was carried out within the framework of the National Recovery and Resilience Plan Greece 2.0, funded by the European Union - NextGenerationEU (Implementation Body: HFRI. Project name: DEEP-REBAYES. HFRI Project Number 15430).}

\author[1]{Iason Chrysomallis}[orcid=0000-0002-1897-321X]
\cormark[1]
\ead{ichrysomallis@tuc.gr}
\credit{Conceptualization, Methodology, Writing -- original draft, Writing -- Review \& editing}

\author[1]{Georgios Chalkiadakis}[orcid=0000-0002-0716-2972]
\ead{gchalkiadakis@tuc.gr}
\credit{Conceptualization, Writing -- Review \& editing}

\affiliation[1]{organization={Technical University of Crete},
                city={Chania},
                postcode={73100}, 
                state={Crete},
                country={Greece}}
                





\cortext[cor1]{Corresponding author}


\begin{abstract}
Imitation learning traditionally requires complete state-action demonstrations from optimal or near-optimal experts. 
These requirements severely limit practical applicability, as many real-world scenarios provide only state observations without corresponding actions, and expert performance is often suboptimal.
In our work in this paper we introduce a complete \textit{deep implicit imitation reinforcement learning} framework that addresses both those limitations by combining deep reinforcement learning with implicit imitation learning from observation-only datasets. 
Our main algorithm,~\textit{Deep Implicit Imitation Q-Network} (DIIQN), employs an action inference mechanism that reconstructs expert actions through online exploration, and integrates a dynamic confidence mechanism that adaptively balances expert-guided and self-directed learning. 
This enables the agent to leverage expert guidance for accelerated training while maintaining the capacity to surpass suboptimal expert performance through environmental interaction.
We further extend our framework with a~\textit{Heterogeneous Actions DIIQN (HA-DIIQN)} algorithm, to tackle scenarios where expert and agent possess {\em different action sets} (or capabilities), a challenge that was previously unaddressed in the implicit imitation learning literature. 
HA-DIIQN introduces an infeasibility detection mechanism and a bridging procedure that identifies alternative pathways connecting agent capabilities to expert guidance when direct action replication is impossible. 
This enables knowledge transfer even when expert transitions are completely infeasible for the agent to execute.
We provide a thorough experimental evaluation of our framework 
across diverse environments.
Our experimental results demonstrate that DIIQN achieves up to $130\%$ higher episodic returns compared to standard DQN, while consistently outperforming existing implicit imitation methods that cannot exceed expert performance. 
In heterogeneous action settings, HA-DIIQN learns up to $64\%$ faster than baselines, successfully leveraging expert datasets that would be unusable by conventional approaches. 
Extensive parameter sensitivity analysis reveals the framework's robustness across varying dataset sizes, and hyperparameter configurations. 
By enabling learning from accessible, suboptimal, observation-only expert data across heterogeneous action spaces, our work expands the practical applicability of imitation learning to real-world scenarios.
\end{abstract}


\maketitle

\section{Introduction}



\textit{Imitation Learning} (IL) refers to the paradigm where a learning agent receives guidance from a proficient agent and attempts to replicate their behaviour. 
The proficient agent is referred to as the expert, while the learning agent is simply called the agent. 
When applied to decision-making problems, IL provides a powerful method for policy transfer to an agent.
This transfer of knowledge may be required for various reasons. 
Initially, IL was developed for cases where the expert model is not always accessible, such as when dealing with human experts.
Later, with the introduction of neural networks, additional use cases emerged, specifically to {\em generalize} the behaviour of predefined restricted mathematical models or algorithms, and to remove the computational overhead associated with them. 
Via imitation learning, we can
infer agent policies
that ``mimick'' the expected expert behaviour, potentially in similar but different settings than the one(s) the expert was observed operating in. 
Moreover, it allows for the learning agent to operate 
efficiently in online settings, which would have been otherwise problematic
due to assumptions or framework limitations.
For instance, complex planning algorithms may require extensive computation time or perfect knowledge of the environment, making them impractical for real-time applications. 
By learning from behaviour demonstrated by such ``expert'' complex algorithms, we can distill their expertise into fast neural network policies that generalize the expert's decision-making patterns without inheriting their computational burden or restrictive requirements.

Now, expert guidance can take many forms. 
In its most common form, it consists of demonstrations, pairs of states and actions that represent what action the expert took at each specific recorded state.
This represents the most intuitive application of imitation learning and falls under the {\em explicit imitation} category, learning from state-action pairs. 
A less common but more practically applicable form of IL is {\em implicit imitation} learning, which only requires {\em observations} 
 instead of {\em demonstrations}. 
In the implicit IL scenario, the expert provides only state transitions, without the associated actions or other information. 
This is important since it significantly simplifies dataset collection.
However, it introduces challenges such as additional algorithmic complexity, as action inference mechanisms must be incorporated to reconstruct the missing action information.
Implicit imitation learning is also known as
\textit{Learning from Observation} (LfO)~\cite{NEURIPS2021_868b7df9,ijcai2018p687,torabi2018generative,liu2024imitation,ijcai2019p882,household,GONZALEZ2022117167}.
In our work, we focus on implicit imitation, leveraging only state observations to guide the learning process; while also addressing an inherent problem of imitation learning, that of the potential suboptimality of the expert.


Indeed, one of the core issues and limitations of imitation learning is the fundamental tendency to learn to mimic the patterns present in the training data. 
If the expert behaviour (and therefore the data provided by the expert) is suboptimal, imitation learning algorithms that do not account for this possibility fall victim to replicating the same problematic ``expert'' behaviours and mistakes. 
Moreover,
without a ground-truth reward or any correcting signal, there is no mechanism to improve beyond the expert's limitations. 
This creates a performance ``ceiling'', that is inherently bounded by the quality of the expert demonstrations, potentially leading to suboptimal policies even with perfect imitation.
Most of the 
current literature implicitly assumes that experts are optimal, or that they at least provide datasets containing predominantly near-optimal behaviours. 
However, this assumption rarely holds in practice. 
Real-world expert data is often suboptimal—despite the terminology, in many practical scenarios we are dealing with fallible demonstrators rather than truly optimal agents. 
This fundamental mismatch between algorithmic assumptions and practical data availability represents a critical limitation of conventional imitation learning approaches. 
In Section~\ref{sec:suboptimal_chall}, we provide an extensive discussion on this suboptimality challenge, and present existing approaches to address it. 

On the other hand, reinforcement learning is a paradigm that, through environmental interaction and reward maximization, pursues optimal policies without requiring any demonstrations. 
Through a procedure of exploration and exploitation, the agent constantly improves its policy by updating its value estimates based on observed rewards and state transitions.
However, this exhaustive exploration constitutes a slow learning procedure, resulting in inefficiently prolonged training that can require millions of environment interactions before achieving satisfactory performance.
The sample inefficiency of pure reinforcement learning approaches often renders them impractical for real-world applications where environmental interaction is costly or time-consuming.


In our work in this paper, 
we bridge these complementary limitations by integrating {\em implicit imitation learning} within {\em an online deep reinforcement learning framework}.
This allows us to benefit from the accurate guidance that provides significantly expedited training compared to regular DRL algorithms, while simultaneously having the leverage and proper feedback from environmental interactions to potentially surpass expert performance when the expert is suboptimal. 
We accomplish this by continuously evaluating our performance alongside the quality of guidance provided by the expert dataset, dynamically choosing whether it is more appropriate to learn from the expert's experience or, when we determine that the expert cannot provide valuable information, to take the lead instead. 
As such, we provide a complete {\em deep implicit imitation reinforcement learning} framework, and an accompanying {\em Deep Implicit Imitation Q-Network (DDIQN)} reinforcement learning.



We initially presented DIIQN's core  in~\cite{chrysomallis2023deep}, and introduced an extension in~\cite{Chrysomallis_Chalkiadakis_Papamichail_Papageorgiou_2025}.
In the latter we study heterogeneous action set scenarios between the expert and the agent. 
This could mean that a human, whether a surgeon or warehouse worker, has much more intricate and detailed actions available than their corresponding robotic counterpart, such as a robotic surgical system or basic warehouse robot, respectively. Alternatively, robots with different degrees of freedom can still communicate and transfer knowledge appropriately when we need to either increase or decrease their capabilities. 
In autonomous driving, vehicles with vastly different acceleration and steering capabilities present another example, as do different simulation scenarios such as varying gravitational fields. 
These applications are abundant and emerge naturally with technological progression, where changes are inevitable, but the requirements to transfer existing behavioural knowledge to newer agent versions are prohibitively expensive.
The real world is heterogeneous---agents naturally possess different capabilities and action sets. 
Humans routinely learn from agents with different capabilities, such as when athletes learn from professionals despite different physical abilities. 
If humans can extract useful behavioural knowledge across such capability gaps, artificial agents should be able to do the same.
Enabling this cross-capability learning is essential for practical applications where knowledge transfer between agents with different embodiments or capabilities is necessary.

\begin{figure}
    \centering
    \begin{subfigure}[b]{0.4\textwidth}
        \centering
        \includegraphics[width=\textwidth]{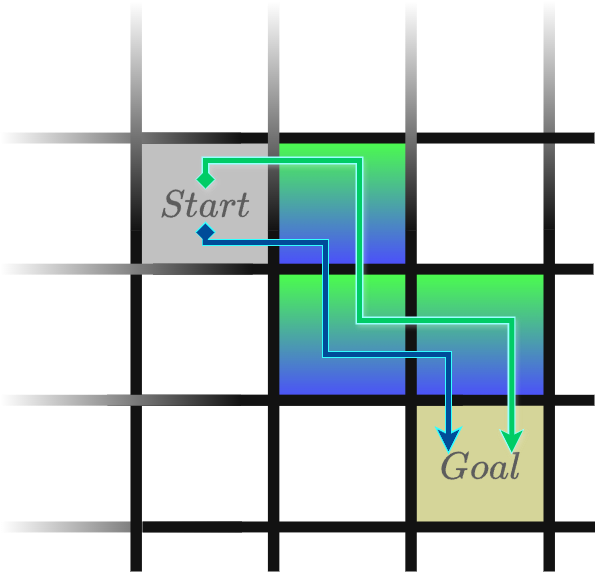}
        \caption{Homogeneous action space}
        \label{fig:heterogeneity_homogeneous}
    \end{subfigure}
    \hspace{0.05\textwidth}
    \begin{subfigure}[b]{0.4\textwidth}
        \centering
        \includegraphics[width=\textwidth]{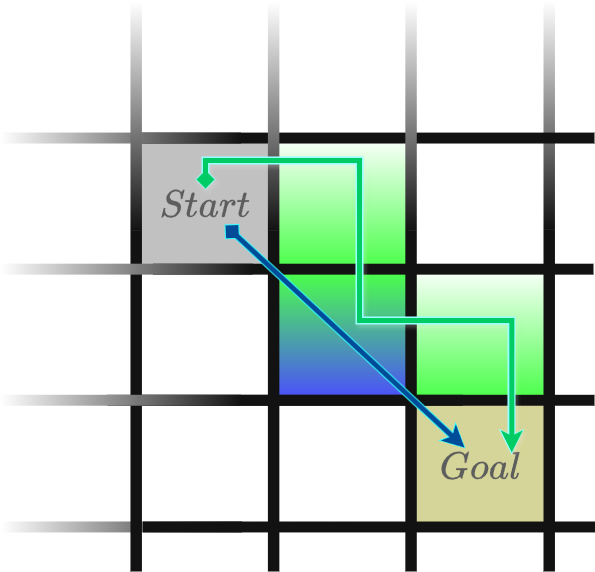}
        \caption{Heterogeneous action space}
        \label{fig:heterogeneity_heterogeneous}
    \end{subfigure}
    \caption{Illustration of action space heterogeneity. (a) Homogeneous scenario where the expert and agent share the same orthogonal actions. (b) Heterogeneous scenario where the agent acts in diagonal actions instead.}
    \label{fig:heterogeneity}
\end{figure}

However, current imitation learning literature focuses predominantly on homogeneous action settings. 
When the expert has different action capabilities than the agent, it is expected that no state transitions provided by the expert can be followed by the agent (Figure~\ref{fig:heterogeneity}). 
This creates a situation where the expert dataset is practically deemed worthless from the IL algorithm's perspective, generating constant confusion and incorrect assumptions during training that lead to ineffective policies.
In our proposed extension, termed~\textit{Heterogeneous Actions Deep Implicit Imitation Q-Network} (HA-DIIQN), we present a method to address this limitation, providing ways to leverage such datasets effectively. 
By examining possible future states, we identify common ground between expert and agent experiences, guiding the agent through these shared state trajectories.

\paragraph{The Challenge of Dataset Accessibility and Expert Optimality}
A substantial portion of the imitation learning literature operates under the assumption that we have complete information from the expert we are learning from. This typically includes the standard state-action pairs~\cite{pomerleau1988alvinn,gail,garg2021iq,kostrikov2019imitation,reddy2019sqil}, or in some cases additional information that provides clarity and guidance for the agent (like preference-based learning methods~\cite{brown2019extrapolating,brown2020better}). It is also commonly assumed that close-to-perfect or fully optimal expert behaviours are represented in the dataset, with no need to employ mechanisms that evaluate their quality. While this approach is intuitive and well-founded, it falls short when confronted with the reality of practical applications.

On one hand, many scenarios present significant barriers to action recording accessibility. 
Complete inability to record actions occurs when observing human behaviour or when actions cannot be parsed through detectors for logging purposes. 
Privacy concerns create another substantial barrier, as many applications permit observing environmental changes while prohibiting individual action logging that could compromise agent privacy. 
Resource constraints further limit feasibility, as obtaining comprehensive action information often proves too expensive for practical dataset creation budgets.

On the other hand, expecting optimal expert performance is also a limiting and unrealistic expectation for two main reasons. 
First, many complex environments lack clearly defined optimal policies, such as in autonomous driving scenarios, making it impossible to define optimal behaviour. 
Consequently, we cannot claim that a dataset believed to operate under a supposedly optimal policy is actually optimal with no room for improvement. 
This leads to creating policies that are fundamentally constrained by unverified assumptions of optimality. 
Second, obtaining non-optimal expert datasets is consistently cheaper and much easier to acquire. Without the need to verify whether a dataset represents optimal expert behaviour, we suddenly enable access to a wealth of potential datasets that can assist a learning agent. 
Naturally, higher quality datasets yield greater benefits for imitation learning approaches.
However, there are many cases where pursuing optimal expert data is simply not worth the investment.

To the best of our knowledge, while both challenges are being individually addressed in the literature, no existing work combines the requirements and constraints of both scenarios to create an implicit imitation learning algorithm that can leverage non-optimal observations and surpass expert performance. To this end, we provide an approach that fills this gap in our research field.

\subsection{Contributions}

In summary, the contributions\footnote{Early versions of this work were presented in~\cite{chrysomallis2023deep,Chrysomallis_Chalkiadakis_Papamichail_Papageorgiou_2025}. 
The current paper provides a unified view of our deep implicit imitation RL framework.
Moreover, it
substantially extends the original framework by modifying the the loss function selection model and introducing a new, 
refined confidence mechanism, that provides more stable dynamic switching between expert guidance and self-directed learning.
This results to both a more theoretically solid approach, and also to improved algorithmic performamce.
Furthermore, the experimental evaluation has been significantly expanded to include multiple new domains, providing deeper insights into the framework's robustness across challenging and diverse environments. For previously evaluated environments, all experiments have been rerun using the updated and optimized framework.
Finally, thorough parameterization studies were conducted as part of our work in this paper.
}
of this work are as follows:
\begin{itemize}
\item We provide a novel deep implicit imitation RL framework, one that gives rise to novel deep implicit imitation RL algorithms and related key components, and allows the agent to surpass suboptimal expert performance. Arguably, ours is the first framework to integrate implicit imitation learning with deep RL: the only other such existing approaches~\cite{kumar2023graph,zhou2024rethinking,swamy2023inverse,gail} resort to devise reward functions to guide the agent learning---usually falling under the {\em inverse RL} research stratum~\cite{arora2021survey}.
\item We present DIIQN, a deep implicit imitation reinforcement learning algorithm that leverages both expert observations and the agent's experience to accelerate training and surpass suboptimal expert performance in environments in which the agent and expert action sets are assumed to be the same.
\item We introduce HA-DIIQN, an extension of DIIQN that handles settings with {\em heterogeneous action sets} between the expert and the agent, enabling the agent to leverage the expert dataset even when step-by-step replication is infeasible.
\item We implement novel, purpose-built for this implicit
imitation domain loss functions, which incorporate guidance from both expert observations and agent experience for homogeneous and heterogeneous action spaces.
\item We develop a confidence mechanism that enables a dynamic combination of expert-guided and self-directed learning, depending on whether or not the agent determines it can surpass expert performance.
In this way the learning agent is able to determining a loss function to optimize against.
\item We provide comprehensive experimental results that demonstrate and validate the proposed algorithm's effectiveness across diverse environments, alongside comparative evaluations against multiple existing implicit imitation learning methods.
\item We conduct extensive parameter sensitivity analysis that offers detailed insights into each algorithmic module's parameterization and establishes recommended ranges for optimal implementation.

\end{itemize}

\subsection{Paper Outline}

The remainder of this paper is structured as follows. 
Section~\ref{sec:background} provides the theoretical foundations and reviews related work in implicit imitation learning, the suboptimality challenge, and deep reinforcement learning techniques. 
Section~\ref{sec:diiqn} presents the Deep Implicit Imitation Q-Network (DIIQN) framework, detailing the core components including action inference, expert sampling, augmented loss functions, and the dynamic confidence mechanism. 
Section~\ref{sec:hetero} extends this framework to heterogeneous action settings, introducing HA-DIIQN with its infeasibility detection and bridge discovery mechanisms. 
Section~\ref{sec:experiments} provides comprehensive experimental evaluation, including comparisons against baseline methods, competing algorithms, and extensive parameter sensitivity analysis. 
Section~\ref{sec:conclusions} concludes with discussion of limitations and future research directions. 
Additional implementation details, distance metric specifications, and hyperparameter configurations are provided in the appendices.

\section{Background and Related Work}
\label{sec:background}

This section establishes the theoretical foundations and reviews related work that informs our DIIQN approach. 
We begin with the foundational concepts of implicit imitation learning, examine the challenges of suboptimal expert data, and provide the necessary background on reinforcement learning techniques that our method builds upon.

\subsection{Implicit Imitation Learning}
Our work is inspired by~\cite{price2003accelerating}, who present a framework designed around the notion of implicit imitation learning that focuses on state-transition-only expert datasets in tabular settings. 
They established the foundations of implicit imitation learning by providing a model-based approach to tackle this challenge, contributing through foundational implicit imitation research and fully defining the problem setting. 
To highlight their assumptions that provide better understanding of the tools required to create an implicit imitation model, we have: 
\textit{Observability}, assuming the agent can fully observe the expert's state without partial observability constraints; 
\textit{Analogy}, assuming both agent and expert operate within identical local state spaces;
\textit{Abilities}, assuming the agent possesses identical action capabilities to the expert, enabling equivalent actions and resulting state transitions; and 
\textit{Objectives}, assuming agent and expert share coinciding goals, receiving equivalent positive and negative reward signals for identical state transitions.
In their experimental evaluation, since this approach was developed for tabular RL, they provide relatively simple but comprehensive experiments in a grid-based two-dimensional maze environment. 

In our
DIIQN
work,
we follow their theoretical foundation and problem assumptions but adapt this paradigm to modern deep learning settings. 
This adaptation enables handling significantly larger state spaces 
with
the introduction of neural networks. 
By leveraging modern DRL techniques, we can accelerate RL agent learning 
via
imitation learning and derive policies for solving complex problems. 
It should be noted that in our HA-DIIQN extension, we break free from the \textit{Abilities} assumption, enabling us to handle heterogeneous action settings scenarios.

\subsection{The Suboptimality Challenge}\label{sec:suboptimal_chall}
Imitation learning has been a field of extensive research interest over the past 25 years.
Starting with the early work of~\cite{pomerleau1988alvinn}, we see the earliest introduction of \textit{Behavioural Cloning} in imitation learning. Subsequently, \textit{Inverse RL}~\cite{ng2000algorithms} (IRL) shifted the goal of imitation learning toward obtaining the expert's reward function instead, while closely dated  works introduced the concept of implicit imitation~\cite{price1999implicit,price2001imitation,price2003accelerating} to tabular approaches. 
Finally, in the last decade, substantial effort has been devoted to transitioning imitation learning to the deep learning era, yielding popular alternative approaches~\cite{ijcai2018p687,gail,torabi2018generative}. 
However, significant challenges remain to be addressed.

Among the challenges of current interest in recent imitation learning research is the topic of suboptimal expert information. 
When we refer to an expert policy as suboptimal, we mean a policy that fails to achieve the maximum possible reward for solving a given problem. 
These scenarios include cases where the expert dataset contains optimal policies but is corrupted with noise, requiring the imitation learning algorithm to filter out the corrupted portions of the given behaviour~\cite{sasaki2021behavioral,yu2024usn,sun2022deterministic,yang2021trail,xu2022discriminator}. 
In such cases, the challenge lies in separating noisy from clean data and learning primarily from the clean examples, rather than leveraging information from all datasets regardless of expertise quality.

A limitation of some of these approaches is their requirement for additional information, such as ranking orders of desired expertise~\cite{brown2019extrapolating,zhou2024rethinking} or manually annotated confidence scores~\cite{pmlr-v97-wu19a}. 
Similarly, in the work of ~\cite{zhang2021confidence}, they deal with mixed datasets that require no labeling, providing a method to derive the best possible policy given the circumstances and offering theoretical guarantees for learning satisfactory behaviour, though without surpassing expert performance. 
To achieve performance beyond expert level, an additional indicating signal is required to provide further insight.

A work 
whose objectives 
are closely aligned with ours,
is
\textit{Disturbance-based Reward Extrapolation} (D-REX)~\cite{brown2020better}, which can handle experts of varying skill levels, automatically rank them, and learn from them to potentially produce superior policies. 
This is accomplished via an IRL approach that initially extracts a reward function fitting their automatic ranking system, and then learns based on that function. 
Like any work aiming to surpass expert performance, it requires access to an online environment to utilize this reward function, thus creating the opportunity to exceed expert performance. 
Under their assumption, if they can understand what the expert dataset aims to accomplish, they can provide an algorithm to exploit it appropriately and potentially deliver better performance. 
However, 
the
D-REX approach
falls under the explicit imitation umbrella, where action information in the dataset is integral to the algorithm's functionality. 
This limitation means it cannot generalize to scenarios where only state observations are available.

\subsection{Markov Decision Processes}
In decision-making scenarios, an agent faces situations requiring action selection to advance and progress within their environment. 
Based on their chosen actions, different outcomes occur, each with negative or positive effects on the agent's desired objectives. 
To formalize this notion, a \textit{Markov Decision Process} (MDP)~\cite{bellman1957markovian,putermanMDPs} provides a mathematical framework designed to describe such problems.
An MDP consists of the tuple $\langle S, A, T, R, \gamma\rangle$. 
The state space $S$ describes snapshots of the current environment with which the agent interacts, while the action space $A$ contains the full set of the agent's capabilities. 
The transition function $T$ determines the next state after the agent takes an action, while simultaneously a reward function $R$ provides positive or negative signals describing the desirability of each state transition outcome. 
Finally, since not all decision-making problems are restricted to single-step rewards (with some benefits hidden deep within specific action sequences) a discount factor $\gamma$ helps prioritize short-term versus long-term rewards through appropriate tuning.

Given this formulation, an agent is tasked with learning a behavioural model to interact with the environment, progress through different states, and receive the maximum possible total reward. 
This behaviour is called the \textit{policy}, which maps states to actions $\pi(s)=a$, meaning that given a current state $s$, it determines the best action $a$ based on the agent's experience. 
An optimal policy $\pi^*(s)$ maximizes the expected cumulative reward, and at least one such policy exists in every MDP. 
Additionally, a \textit{value function}~\cite{sutton1998reinforcement} can evaluate the agent's current position to determine whether the current state is favorable in terms of expected future rewards.

\subsection{Deep Reinforcement Learning}

In order to create a policy for problems defined within an MDP framework, when we lack prior knowledge of effective actions and/or policies, one approach is through \textit{Reinforcement Learning} (RL)~\cite{sutton1998reinforcement}. 
In these cases, the RL agent can interact with the environment, and while the reward function is not directly accessible, it is environmentally available, meaning the agent receives appropriate reward signals after each action taken. 
Through a procedure of exploration and exploitation, an RL agent strategically switches between randomly exploring and following its own intuitions to achieve progressively higher rewards.

Combining this technique with neural networks leads to \textit{Deep Reinforcement Learning} (DRL). 
A critical advantage of DRL lies in its ability to handle high-dimensional state spaces through function approximation, enabling learning in complex environments where tabular methods become computationally infeasible.
Arguably the most well-known DRL algorithm is the \textit{Deep Q-Network} (DQN)~\cite{DQN}, which enables numerous new techniques that help transition the RL framework to deep learning. The agent policy is a neural network $Q(s,a;\theta)$ operating under parameters $\theta$ that, given a state $s$ and chosen action $a$, provides the expected cumulative reward. 
To train this model, a target network using past parameters $\theta^-$ is required: $y=r+\gamma max_{a'} Q(s',a';\theta^-)$. By calculating the temporal difference (TD), the loss function is computed and minimized for learning and progression:
\begin{equation}
    L(\theta)=E_{(s,a,r,s')} D[(Q(s,a;\theta) - y)^2]
\end{equation}
The target network provides stability by preventing targets from changing too rapidly during training. 
To gather training data, \textit{replay memory} stores all MDP-related information from every step as samples in a queue. 
Training in batches constructed through uniformly random sample selection ensures that consecutive state chains cannot create policy bias. 
On top of that, several subsequent improvements have emerged and are now integral to the algorithm's best version.

Double Deep Q-Network (DDQN)~\cite{van2016deep} addresses overestimation bias inherent in DQN by decoupling action selection from action evaluation. 
While DQN uses the target network for both selecting and evaluating the best next action, DDQN uses the main network for action selection. 
Since the network selecting actions differs from the one evaluating them, this prevents overly optimistic Q-value estimates. 
This modification changes the target function to:
\begin{equation}
y^{DDQN}=r+\gamma Q\left(s'\operatorname*{argmax}_{a'}Q(s',a';\theta);\theta^-\right)
\end{equation}

Prioritized Experience Replay (PER)~\cite{schaul2015prioritized} enhances the replay memory module of DQN for improved batch sampling. 
Instead of uniformly selecting training samples at each step, a stochastic selection favors samples with higher TD errors. 
This follows the principle that higher TD differences indicate larger gaps requiring policy improvement. 
To avoid bias from non-uniform sampling, importance sampling weights~\cite{mahmood2014weighted} are incorporated. 
By focusing on the most critical experiences, this approach leads to faster convergence and better guidance for the DRL agent.


\section{Deep Implicit Imitation Reinforcement Learning}
\label{sec:diiqn}

In this section, we present 
our
deep implicit reinforcement learning framework, a paradigm that integrates implicit imitation learning within deep RL to enable learning from observation-only expert datasets, while maintaining the capacity to surpass suboptimal expert performance. 
We present 
all
components of this framework and demonstrate how their integration gives rise to DIIQN, the algorithm that realizes this framework by adaptively balancing expert guidance with self-directed learning.

In the proposed framework, DIIQN is a DRL-based technique that leverages implicit imitation learning. 
At its core, it follows the training steps of DQN, meaning that we use $\epsilon$-greedy exploration to traverse through the state space and interact with the environment, keeping transition samples in the replay memory that we will later use to train our model. 
The imitation learning component is enabled at loss function computation, where we integrate the guidance of the expert. 
We go step by step through what is required for the implicit imitation learning model to be activated within a DQN execution, providing us the DIIQN method. 
We begin by explaining how the {\em  collection of expert data} that consists of only states is performed; then, we detail {\em how to infer the actions} associated with each expert transition; afterwards, we explain {\em how to sample} from the expert dataset so that we can train appropriately by {\em introducing our augmented loss functions} required to facilitate implicit imitation learning. 
Finally, we describe {\em the mechanism that enables dynamic switching between 
these loss function},
based on the judgment of the agent---intuitively, on whether it decides to keep the influence of the expert high if it can still learn from it, or low, if the expert is deemed of no further use.

Subsequently, in Section~\ref{sec:hetero}, we extend DIIQN within the deep implicit imitation reinforcement learning framework to accommodate heterogeneous action spaces, introducing HA-DIIQN to handle scenarios where expert and agent possess fundamentally different action capabilities.

\subsection{Data Collection}

Since we are working with an implicit imitation learning framework, the expert provides only state information without corresponding actions. 
This expert data can be structured in two possible formats: sequential states $\{s_e^{(1)}, s_e^{(2)}, \ldots, s_e^{(N)}\}$ collected in order of execution, or state transition pairs $\{(s_e, s_e')\}$ that preserve the temporal relationships between consecutive states. 
The latter format is particularly useful when sequential ordering cannot be guaranteed, as it maintains the necessary state correlations for learning.

The number of state transitions required varies depending on the complexity of the environment and the task that the agent is trying to solve. 
We use behavioural cloning as an illustrative example to explain why imitation learning generally requires large datasets.
In behavioural cloning scenarios, despite being categorized as a supervised learning algorithm for decision making, substantially more data is typically required compared to standard supervised learning problems~\cite{belkhale2023data}. 
In supervised learning, samples are independent and identically distributed (i.i.d.), requiring only an output for each individual sample. 
In contrast, decision-making problems often require a chain of actions executed in a specific sequence, and constructing these behavioural chains demands significantly more data. 
In our case, however, since DIIQN leverages agent experience to fill in the gaps and can even develop policies that surpass expert performance, datasets of any size can benefit the agent. 
Large datasets are expected to provide more comprehensive guidance across different scenarios, while smaller datasets will result in slower learning but still yield meaningful results---a scenario where other imitation learning techniques would typically struggle.

Any form of policy can generate the sequence of expert states, ranging from human operators performing tasks to mathematical functions or intelligent agents. 
In our research, we employed the latter approach, specifically using agents trained with DQN---the baseline DRL algorithm upon which we build our framework. 
While several datasets from different behavioural models are available for popular environments online~\cite{minari}, providing rich and diverse state transitions, we created our own datasets for our experiments.

Once the observation dataset is provided, the expert plays no further role in the agent's training procedure. 
The agent imports this information and proceeds to process the observations appropriately to learn an effective policy.

\paragraph{Multiple Experts}

The dataset can incorporate observations and experiences from different experts by merging them into a large collection.
However, this aggregation should be done carefully, as its effectiveness depends on the environment and the quality of the experts.
On one hand, incorporating multiple experts provides valuable diversity, which typically benefits imitation learning algorithms by enabling more active learning during training and accelerating convergence. 
This is particularly important in implicit imitation scenarios where environment interactions are utilized, as seen in methods like~\cite{ijcai2018p687,torabi2018generative}. 
Conversely, when experts exhibit contradictory behaviours or vary significantly in their levels of optimality, the resulting mixed signals can confuse the learning agent. 
While this challenge is not unique to our approach but rather an inherent limitation of imitation learning in general~\cite{gandhi2023eliciting}, our DIIQN framework demonstrates robustness in handling such contradictions by leveraging the agent's own experience to resolve ambiguities and develop coherent policies independently.
Our experimental evaluation employs multi-expert datasets throughout (see Section~\ref{sec:exper_setup}), validating the framework's ability to extract valuable guidance while mitigating the effects of expert inconsistencies.

\subsection{Action Inference}
\label{action_inference}

After importing the expert dataset, the agent faces its first challenge in the implicit imitation scenario: inferring the actions associated with the observed state transitions. 
Since no signal from the expert or any other offline data can guide the agent in understanding the underlying dynamics that produced the expert's state transitions, we must rely on the available online learning tools.

To fill this information gap, we utilize the agent's online interactions with the environment. 
Following the standard DQN paradigm, the agent transitions from exploration to exploitation while maintaining a record of its experiences in a replay buffer. 
These experiences $\langle s_a, a_a, s_a' \rangle$ contain complete information about the agent's state transitions $\langle s_a, s_a' \rangle$ and the corresponding actions $a_a$ taken by the agent.

We utilize these experiences not only for optimizing the agent's behaviour, but also for updating the inferred actions of the expert dataset. 
For each agent-discovered transition $\langle s_a, a_a, s_a' \rangle$, we compute a divergence using a distance metric $D$ between this transition and each expert transition $\langle s_e, s_e' \rangle$. 
This divergence represents how closely the current transition matches an expert transition and can be interpreted as an error metric indicating how far the agent-discovered action $a_a$ potentially deviates from the actual expert action $a_e^{real}$ taken in the expert transition.
For each newly explored transition, if the computed divergence is lower than the one currently assigned to the expert transition, this indicates that the current action provides a better estimation for the expert's inferred action. 
Consequently, we update both the inferred action and the divergence metric. 
This divergence metric is referred to as the \textit{action error metric}.

Regarding the definition of the distance metric, the choice depends on the specific environment characteristics. 
We cannot apply the same distance metric to an environment with a two-dimensional state space as we would to an environment that provides full-screen pixel observations. 
In our experiments, this distance metric ranges from Euclidean distance in simpler environments to weighted Hamming distance in more complex environments. 
For detailed information on distance metric selection, see Appendix~\ref{apdx:distances}.

To avoid starting with completely unreliable action estimates, a cold start procedure can be optionally enabled, similar to~\cite{ijcai2018p687}. 
Cold start refers to a pre-training period where the agent executes random actions purely for exploration to populate initial experience. 
In our case, this procedure provides initial action estimations, particularly for early state transitions, establishing a stable foundation upon which to build expert information. 
This approach is also commonly used in DQN~\cite{DQN} applications to increase the number of samples in the replay memory, ensuring efficient random batch sampling from the beginning.

\subsection{Expert Sampling}
\label{sec:expert_sampling}

After establishing a method to infer expert actions, we proceed to explain how to exploit the expert dataset. 
The first step in training is to appropriately sample from the expert dataset. 
Our approach involves storing expert transitions alongside the agent's transitions so that both can be utilized by our loss functions. 
This means that for each sample taken from online environment interaction during normal DQN exploration or exploitation, we also store a relevant transition from the expert dataset.

\begin{figure}
  \centering
\includegraphics[width=0.4\textwidth]{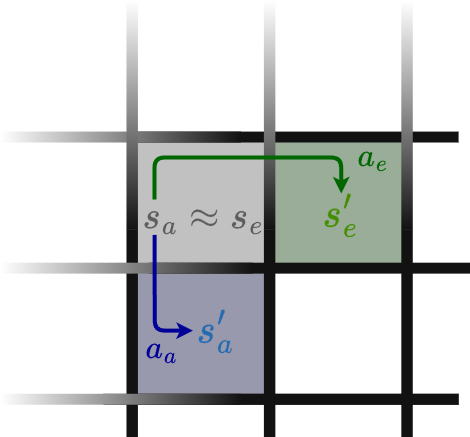}
\caption{Starting from a similar initial state between the agent and the expert ($s_a \approx s_e$), the trajectories diverge as each takes a different action.
The agent executes action $a_a$, leading to the transition $s_a \rightarrow s_a'$, while the expert performs action $a_e$, resulting in the transition $s_e \rightarrow s_e'$.
}
\label{fig:sampling}
\end{figure}

This relevant transition represents what the expert would have done in the agent's position. 
Since we have a collection of expert state transitions and their inferred actions, for any given agent transition $\langle s_a, a_a, s_a' \rangle$, we can find an expert transition $\langle s_e, a_e, s_e' \rangle$ with a similar initial state $s_a \approx s_e$. 
This expert transition represents the action sequence the expert followed when faced with the same initial state, rather than what the agent chose (Figure~\ref{fig:sampling}). 
In cases where the agent can benefit from expert guidance, we expect the expert transition to differ from the agent's chosen path.

For the search procedure to find an appropriate expert transition, we execute a k-nearest neighbor (KNN) search~\cite{cover1967nearest,arya1998optimal} using only the initial state of each transition. 
This yields $k$ expert transitions with starting states $s_e$ that are closest to our current agent state $s_a$. 
The distance metric used in this search procedure is the same as the one employed in the \textit{Action Inference} phase, which depends on the specific application.

\begin{figure}
  \centering
\includegraphics[width=0.55\textwidth]{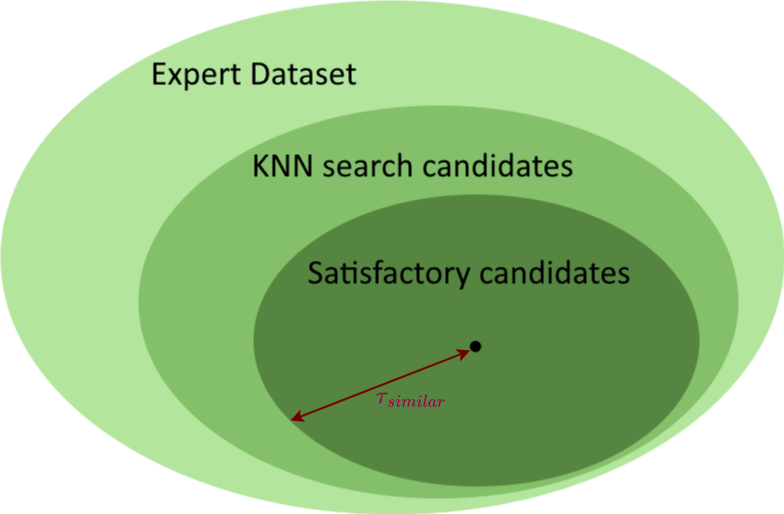}
\caption{Expert sampling procedure with KNN search and similarity filtering. The expert dataset (outer region) contains all available expert transitions. A KNN search identifies the $k$ nearest expert states to the current agent state (middle region). From these candidates, only those within the similarity threshold $\tau_{similar}$ are retained as satisfactory candidates (inner region). The arrow illustrates the maximum acceptable distance $\tau_{similar}$ between states for them to be considered similar. One expert transition is randomly selected from the satisfactory candidates to provide guidance for the current training step. If no candidates satisfy the similarity threshold, expert guidance is skipped for that step.}
\label{fig:knn_search}
\end{figure}

However, we have not yet defined what constitutes ``similar'' states. 
Simply obtaining the nearest neighbors does not guarantee that the returned transitions will have truly similar initial states. 
For this reason, from the batch returned by the KNN search, we only retain candidates that are similar within a certain threshold $\tau_{similar}$ (Figure~\ref{fig:knn_search}).
This is hard constraint, meaning that if no candidates fit the criteria, we skip the expert guidance procedure altogether.
We recommend normalizing the state space for this part of the algorithm, allowing the threshold to be expressed as a percentage. 
Only when the state space is normalized can we determine the upper and lower bounds necessary for a percentile estimation of state similarity. 
For detailed information on normalization approaches, see Appendix~\ref{apdx:distances}. 
Alternatively, if state normalization is not feasible, empirical thresholds must be used to define satisfactory similarity levels for accepting candidate states.
After filtering the candidates to obtain the final group of satisfactory states, we randomly select one of them, ensuring that no bias is introduced into our learning procedure.

As a result of this procedure, at each time step, the agent stores a sample in the replay memory that includes $\langle s_a, a_a, s_a', r, s_e, a_e, s_e' \rangle$. 
We refer to this as an \textit{augmented experience tuple}, which contains information about the current agent transition $\langle s_a, a_a, s_a' \rangle$, the reward $r$, and the corresponding expert transition with its inferred action $\langle s_e, a_e, s_e' \rangle$, where $s_a \approx s_e$.

Algorithmically, \textit{Action Inference} and \textit{Expert Sampling} can be performed simultaneously. 
During the neighbor search procedure, while computing distances to establish the neighbor ordering, the corresponding action and error estimations can be updated to appropriate values. 
If the nearest neighbor search is implemented through an optimized library that does not allow interference to update corresponding values, or if comparing the entire expert dataset is computationally expensive and time-consuming, one can increase the nearest neighbor parameter to return a larger portion of the closest state transitions and update only their values rather than the entire dataset.

\subsection{A Confidence Mechanism to Compute the Final Loss Function}
\label{sec:diiqn-loss}

Given samples in the form of $\langle s_a, a_a, s_a', r, s_e, a_e, s_e' \rangle$, the agent is now tasked with training its neural network model using the information from each sample. 
Similar to the standard DQN loss function, we introduce two loss functions: one incorporating information from the agent's collected transition $\langle s_a, a_a, s_a' \rangle$, denoted as $L(\theta_t)_a$, and another incorporating information from the expert's state transition with inferred action $\langle s_e, a_e, s_e' \rangle$, denoted as $L(\theta_t)_e$. 
Their complete forms are as follows:

\begin{equation}\label{augm_obs}
    L(\theta_t)_a = \mathbb{E}_{s,a,r,s'}\left[(r + \gamma Q(s'_a, \operatorname*{argmax}_{a'} Q(s'_a, a'; \theta_t^-); \theta^-_t) - Q(s_a, a_a; \theta_t))^2\right]
\end{equation}

\begin{equation}\label{augm_ment}
    L(\theta_t)_e = \mathbb{E}_{s,a,r,s'}\left[(r + \gamma Q(s'_e, \operatorname*{argmax}_{a'} Q(s'_e, a'; \theta_t^-); \theta^-_t) - Q(s_e, a_e; \theta_t))^2\right]
\end{equation}

However, having two separate loss functions, one for learning from agent experiences and one for learning from expert information, creates the need to determine how to balance learning from both sources. 
A mechanism is required that can switch between expert guidance and self-directed learning when we determine that the agent has extracted sufficient knowledge from the expert and should explore potentially superior behaviours. 
To address this, we introduce a novel \textit{confidence mechanism} that makes this decision dynamically during training.

This confidence mechanism must consider multiple factors. It needs to: a) \textbf{assess} how good the expert action is based on our current understanding of Q-values, b) \textbf{discount} this estimation by computing our confidence in the accuracy of these Q-values for those states, and c) \textbf{adjust} the result based on the inferred action error metric.

To \textbf{assess} whether an expert action is beneficial to focus on and learn from, we apply our Q-model to the initial expert state and examine the divergence between Q-values for the agent action $Q(s_e, a_a)$ and the expert action $Q(s_e, a_e)$. 
This divergence is then passed through a sigmoid function to constrain the value between $(0,1)$, providing us with:
\begin{equation}
    \Delta Q(s_e,a_e,a_a) = \sigma\left(\left[Q(s_e, a_e) - Q(s_e, a_a)\right] \cdot \beta\right)
\end{equation}
where $\beta$ is a hyperparameter that controls the sharpness of the sigmoid function. 
This provides a calculated measure of how much better the expert's action choice is compared to the agent's chosen action at that step. 
Higher values indicate beneficial expert behaviour that the agent should learn from, while lower values suggest that the agent should not follow the expert's behaviour closely, as it can potentially outperform it.

However, the above approach is not always reliable since it depends on the current Q-value model. 
If the model has not been sufficiently trained, we cannot expect it to provide accurate, global estimates of the Q-function that we can rely on to determine which loss function to learn from. 
For this reason, we also employ a \textbf{discount} weight designed to reduce the divergence by computing how frequently the model has trained on the region around the expert state.
We maintain a counter $c_{s_e}$ for each state $s_e$ that increments by one whenever that expert state is selected as part of a training update. 
We also increment this counter when the state $s_e$ appears in the final pool of similar states but is not chosen due to random selection. 
This is justified because if the local region around the expert state has been sufficiently trained to provide reliable Q-values, we expect that the expert state would also have good estimates.
Additionally, we define a parameter $c_{max}$ that represents the desired number of training updates for each expert state region. 
Since training may exceed this threshold, we constrain the counter $c_{s_e}$ to this maximum value to ensure the weight remains bounded at $1$, as this serves purely as a discount factor.
By defining this empirically determined maximum value $c_{max}$, we can formulate the discount weight as:
\begin{equation}\label{disc_weight}
    w(s_e) = \frac{\log(1+\min(c_{s_e}, c_{max}))}{\log(1+c_{max})}
\end{equation}
This value is bounded in $[0,1]$, where higher values indicate greater trust in the Q-model estimates, while lower values indicate the opposite.

The final component addresses the reliability of the action inference. 
Since we already have the action error metric $err_{a_e}$ for each expert sample (a distance metric representing the error of the inferred action), we can derive the last component of the confidence mechanism. 
For the recommended cases with normalized state space, we can obtain a maximum possible error $err_{max}$ and compute the following normalized error ratio to \textbf{adjust} for inference reliability:
\begin{equation}\label{inference_rel}
    \epsilon(s_e) = 1 - \frac{err_{a_e}}{err_{max}}
\end{equation}
This value is also bounded in $[0,1]$, where high values indicate reliable action inference and low values indicate unreliable inference. 
For detailed information about computing the maximum error, see Appendix~\ref{apdx:distances}.

Now that all three components have been defined, the final confidence mechanism is:
\begin{equation}
   \Phi(s_e,a_e,a_a) = \min\left(\Delta Q(s_e,a_e,a_a) \cdot w(s_e), \epsilon(s_e)\right)
\end{equation}
In summary, we compute the value of the expert state with $\Delta Q(s_e,a_e,a_a)$, discount it based on our trust in this evaluation using $w(s_e)$, and finally clip the result based on $\epsilon(s_e)$. 
When the action inference is reliable, the ceiling imposed by $\epsilon(s_e)$ for trusting the expert is high, allowing significant expert influence. 
Conversely, when action inference is unreliable, it prevents expert influence from affecting the learning process.
The characteristics and interpretations of all components are summarized in Table~\ref{tab:confidence}.

With all components in place, the final loss function is:
\begin{equation}
    L(\theta_t) = \Phi(s_e,a_e,a_a) L(\theta_t)_e + \left(1-\Phi(s_e,a_e,a_a)\right) L(\theta_t)_a
\end{equation}
This combines both agent and expert loss functions in a dynamic manner, adaptively adjusting the relative contributions based on the confidence mechanism.

It is important to note that the confidence mechanism provided by our implicit imitation RL framework can be incorporated into {\em any} other deep implicit imitation RL algorithm beyond DIIQN---i.e., the underlying deep RL algorithm of choice does not have to be DQN.
Depending on the underlying algorithm, alternative assessment methods could replace Q-value divergence---for instance, value function estimates in actor-critic methods, policy probability ratios, or learned discriminators in adversarial approaches. 
At its core, the confidence mechanism serves as a {\em general-purpose} dynamic weighting component that adaptively balances expert influence against self-directed learning based on the reliability and quality of available guidance. 
Furthermore, while our implementation focuses on discrete action spaces, the core principles  extend naturally to continuous domains. 
For continuous state spaces, training frequency can be assessed by measuring local density of visited states rather than exact counts~\cite{replay_graph}. Similarly, action inference confidence in continuous action spaces can be estimated through standard regression uncertainty measures that quantify prediction reliability~\cite{yao2019negative}.

\begin{tabularx}{\textwidth} { 
   >{\centering\arraybackslash}m{2cm} 
   >{\centering\arraybackslash}m{2.5cm} 
   >{\centering\arraybackslash}m{2.5cm} 
   >{\centering\arraybackslash}m{1cm} 
   >{\centering\arraybackslash}m{2.5cm} 
   >{\centering\arraybackslash}m{2.5cm} }
    \caption{DIIQN Confidence Mechanism Components and Their Characteristics}\label{tab:confidence}\\
    \hline
    Symbol & Description & Purpose & Range & Low Value & High Value \\
    \hline
    \hline
     $\Delta Q$ & Q-value divergence & Expert quality assessment & $(0,1)$ & Agent outperforms expert & Expert outperforms agent \\
    \hline
     $w(s_e)$ & Training frequency weight & Q-model reliability & $[0,1]$ & Undertrained state region & Well-trained state region \\
    \hline
     $\epsilon(s_e)$ & Action inference confidence & Inference reliability & $[0,1]$ & Poor action inference & Accurate action inference \\
    \hline
     $\Phi$ & Overall confidence & Final expert influence & $[0,1]$ & Ignore expert guidance & Follow expert guidance \\
    \hline
\end{tabularx}

\subsection{A novel deep implicit imitation RL algorithm}
In the previous sections we presented the key components of our framework in detail.
We determined how to create an expert dataset; we explained how to infer expert actions; we described how to sample from the expert dataset; and, finally, we detailed how to train with this information.
Putting these all together, gives rise to our proposed deep implicit imitation RL algorithm, DIIQN, that can be employed in settings with homogeneous actions spaces. The algorithm is presented in  Algorithm~\ref{alg:diiqn}, which details the full DIIQN training loop, illustrating the connection between the aforementioned components during agent learning.
In the next section, we will detail how  to extend DIIQN to heterogeneous action settings as well.

\begin{algorithm}[H]
\caption{DIIQN Training Loop}
\label{alg:diiqn}
\begin{algorithmic}[1]
\REQUIRE Expert dataset $\mathcal{D}_e = \{(s_e^{(i)}, s_e'^{(i)})\}_{i=1}^N$, environment $\mathcal{E}$
\REQUIRE Hyperparameters: $\gamma$, $\beta$, $c_{max}$, $\tau_{similar}$, batch size $B$, $k$ (KNN parameter)

\STATE Initialize Q-network $Q(s,a;\theta)$ and target network $Q(s,a;\theta^-)$ with random weights
\STATE Initialize replay memory $\mathcal{D}_a = \emptyset$
\STATE Initialize expert action inferences: $a_e^{(i)} \leftarrow \text{random}$, $err_{a_e}^{(i)} \leftarrow \infty$ for all $i$
\STATE Initialize expert state counters: $c_{s_e}^{(i)} \leftarrow 0$ for all $i$
\STATE \textbf{Optional:} Cold start - populate $\mathcal{D}_a$ with random exploration

\FOR{episode $= 1$ to $M$}
    \STATE Initialize state $s_a \leftarrow \mathcal{E}.\text{reset}()$
    
    \WHILE{not terminal}
        \STATE Select action $a_a$ using $\epsilon$-greedy policy based on $Q(s_a, \cdot; \theta)$
        \STATE Execute $a_a$, observe reward $r$ and next state $s_a'$
        
        \STATE \textcolor{blue}{// \textbf{Action Inference (Section 3.2)}}
        \FOR{each expert transition $(s_e^{(i)}, s_e'^{(i)})$ in $\mathcal{D}_e$}
            \STATE Compute divergence: $d = D((s_a, a_a, s_a'), (s_e^{(i)}, s_e'^{(i)}))$
            \IF{$d < err_{a_e}^{(i)}$}
                \STATE $a_e^{(i)} \leftarrow a_a$ \COMMENT{Update inferred action}
                \STATE $err_{a_e}^{(i)} \leftarrow d$ \COMMENT{Update action error}
            \ENDIF
        \ENDFOR
        
        \STATE \textcolor{blue}{// \textbf{Expert Sampling (Section 3.3)}}
        \STATE Find $k$-nearest expert states: $\mathcal{N}_k \leftarrow \text{KNN}(s_a, \{s_e^{(i)}\}, k)$
        \STATE Filter by similarity: $\mathcal{N}_{filtered} \leftarrow \{s_e \in \mathcal{N}_k : D(s_a, s_e) \leq \tau_{similar}\}$
        \IF{$\mathcal{N}_{filtered} \neq \emptyset$}
            \STATE Randomly select $s_e$ from $\mathcal{N}_{filtered}$ with corresponding $a_e, s_e'$
            \STATE Store augmented sample: $\mathcal{D}_a \leftarrow \mathcal{D}_a \cup \{(s_a, a_a, s_a', r, s_e, a_e, s_e')\}$
            \algstore{myalg}
\end{algorithmic}
\end{algorithm}

\begin{algorithm}[H]
\begin{algorithmic}[1]                
\algrestore{myalg}
        \ELSE
            \STATE Store standard sample: $\mathcal{D}_a \leftarrow \mathcal{D}_a \cup \{(s_a, a_a, s_a', r, \text{null})\}$
        \ENDIF
        
        \STATE $s_a \leftarrow s_a'$
        
        \STATE \textcolor{blue}{// \textbf{Training Step}}
        \IF{time to update}
            \STATE Sample minibatch of $B$ samples from $\mathcal{D}_a$
            
            \FOR{each sample $(s_a, a_a, s_a', r, s_e, a_e, s_e')$ in minibatch}
                \STATE \textcolor{blue}{// \textbf{Loss Function Computation (Section 3.4)}}
                \STATE Compute agent loss (Eq. 3):
                \STATE \quad $L(\theta)_a = \left(r + \gamma Q(s_a', \arg\max_{a'} Q(s_a', a'; \theta^-); \theta^-) - Q(s_a, a_a; \theta)\right)^2$
                
                \IF{expert transition available}
                    \STATE Increment counter: $c_{s_e} \leftarrow \min(c_{s_e} + 1, c_{max})$
                    
                    \STATE Compute expert loss (Eq. 4):
                    \STATE \quad $L(\theta)_e = \left(r + \gamma Q(s_e', \arg\max_{a'} Q(s_e', a'; \theta^-); \theta^-) - Q(s_e, a_e; \theta)\right)^2$

                    \STATE \textcolor{blue}{// Confidence Mechanism Components}
                    \STATE Q-value divergence (Eq. 5): $\Delta Q(s_e, a_e, a_a) = \sigma([Q(s_e, a_e) - Q(s_e, a_a)] \cdot \beta)$
                    \STATE Training frequency weight (Eq. 6): $w(s_e) = \frac{\log(1 + c_{s_e})}{\log(1 + c_{max})}$
                    \STATE Action inference confidence (Eq. 7): $\epsilon(s_e) = 1 - \frac{err_{a_e}}{err_{max}}$
                    
                    \STATE Final confidence (Eq. 8): $\Phi(s_e, a_e, a_a) = \min(\Delta Q(s_e, a_e, a_a) \cdot w(s_e), \epsilon(s_e))$
                    
                    \STATE Combined loss (Eq. 9): $L(\theta) = \Phi \cdot L(\theta)_e + (1 - \Phi) \cdot L(\theta)_a$
                \ELSE
                    \STATE $L(\theta) = L(\theta)_a$ \COMMENT{Use only agent loss}
                \ENDIF
            \ENDFOR
            
            \STATE Perform gradient descent step on $\sum L(\theta)$ to update $\theta$
        \ENDIF
        
        \IF{time to update target network}
            \STATE $\theta^- \leftarrow \theta$
        \ENDIF
    \ENDWHILE
\ENDFOR
\end{algorithmic}
\end{algorithm}


\section{Heterogeneous Actions DIIQN}
\label{sec:hetero}
Proceeding to the second component of our methodology, we focus on heterogeneous action settings by introducing HA-DIIQN, an extension within the deep implicit imitation reinforcement learning framework. 
It is important to note that heterogeneous action sets can occur in two distinct scenarios. 
The first involves partial overlap, where the agent shares a subset of its action repertoire with the expert, enabling execution of certain expert movements. 
The second scenario involves complete incompatibility, where the agent and expert possess no overlapping action capabilities, rendering all expert transitions infeasible for the agent.

In both cases, the agent lacks prior knowledge regarding which portions of the action set are infeasible and to what extent. 
Consequently, the algorithm must handle both semi-overlapping and completely disjoint action settings uniformly. 
For the implementation of HA-DIIQN, we begin by defining transition infeasibility and establishing detection mechanisms. 
Following this, we explain how to manage infeasible transitions by discovering alternative pathways that preserve expert guidance consistency and maintain fundamental behavioural reasoning.
Finally, we demonstrate how information derived from these pathways is incorporated into the learning process to enhance policy development.
Beyond these novel components, the remainder of the model closely follows the DIIQN architecture, maintaining identical exploration strategies and high-level training principles.

\subsection{Infeasibility Identification}

The initial challenge involves determining which expert transitions are infeasible given the agent's current capabilities and constraints. 
This requires establishing a classification mechanism that distinguishes between feasible and infeasible expert transitions. 
Since the agent lacks prior knowledge regarding the disparities in action space dynamics between itself and the expert, this identification process must be conducted dynamically during training as the agent acquires understanding of the environmental action dynamics.

During training, at each step, the agent executes the standard DIIQN cycle, which includes sampling an expert transition $\langle s_e, a_e, s_e' \rangle$. 
We recall that, due to the implicit imitation nature of the algorithm, the state transition is provided by the expert while the action is inferred by the agent. 
Each inferred action $a_e$ has an associated action error metric $err_{a_e}$ that quantifies the deviation between the inferred action and the actual expert action, which remains unknown to us (see Section~\ref{action_inference}).
This error metric serves as an indicator of expert transition infeasibility. 
High error values suggest that the current action estimation is inaccurate, indicating a high probability that the transition represents an infeasible action sequence given the agent's capabilities. 
Consequently, such transitions are classified as infeasible.

We acknowledge that high action error estimation does not necessarily indicate transition infeasibility. 
It may alternatively signify that a similar transition has not yet been explored by the agent, preventing accurate action inference. 
In such cases, this transition would provide no value to the agent, as it would not pass the KNN search due to its high error evaluation. 
Therefore, labeling it as infeasible provides an alternative mechanism for leveraging this transition.
Importantly, this estimation undergoes continuous updates as the agent explores the environment's state space. 
For this reason, there is a high chance that as the agent develops its policy and achieves improved performance, the transition will be populated with accurate estimations, dramatically reducing the error and reclassifying it from infeasible to feasible.

Regarding the constraint that determines transition infeasibility using the action error estimation as a metric, in the recommended case of normalized state space, a percentile can be employed as the error infeasibility threshold. 
By maintaining this threshold at moderately high levels (e.g., $80$th percentile), we ensure that the agent characterizes the majority of transitions with low confidence as infeasible.

%

\subsection{Bridging the Infeasibility}

Labeling a transition as infeasible is insufficient for extracting learning value from it. 
A straightforward approach would be to simply exclude these transitions from the agent's learning procedure. 
In favorable scenarios, this would result in a substantial reduction of the expert dataset, particularly when the agent shares only a portion of its action space with the expert. 
In adverse scenarios, where the agent possesses completely different action capabilities from the expert, this would effectively render all expert observations unusable for learning purposes. 
Therefore, we must establish a methodology that exploits these expert transitions despite the agent's inability to replicate them precisely.

\begin{figure}
  \centering
\includegraphics[width=0.7\textwidth]{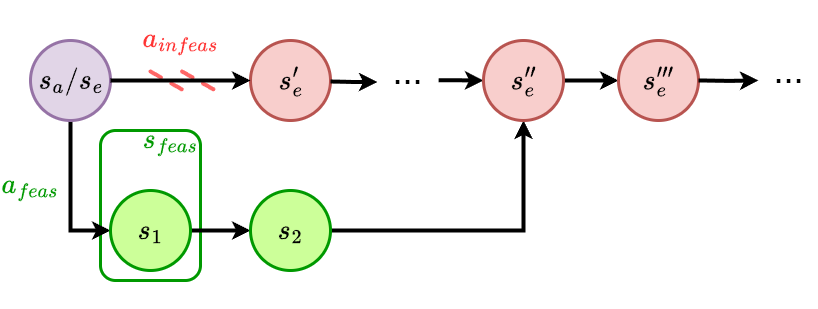}
\caption{Illustration of bridge discovery in heterogeneous action spaces. Starting from similar states $s_a \approx s_e$, the expert executes infeasible action $a_{infeas}$  to transition to $s_e'$, which the agent cannot replicate. Instead, the agent discovers a feasible bridge path via $s_a \xrightarrow{a_{feas}} s_1 = s_{feas} \rightarrow s_2$ that intersects with the expert's downstream trajectory at state $s_e''$. This bridge enables the agent to align with the expert's strategic guidance despite the action space heterogeneity.}
\label{fig:infeas_trajectory}
\end{figure}


It is reasonable to assume that while the agent cannot immediately reach the next expert state $s_e'$ with a single action, it may potentially achieve this position through a sequence of actions. 
Alternatively, the agent may not reach that exact state but could arrive at a subsequent state in the expert's trajectory (Figure~\ref{fig:infeas_trajectory}). 
In other words, there may exist an action sequence that the agent can execute which intersects with the expert's state trajectory.
Although the agent cannot replicate the expert's actions step-by-step, it can discover alternative pathways that align with the expert's suggested guidance. 
We term these alternative pathways \textit{bridges}, as they bridge the gap between agent and expert capabilities. 
To construct these bridges, two sets of trajectories must be considered: the feasible paths available to the agent and the projected paths of the expert.

Determining expert trajectories is relatively straightforward. 
In the standard case where expert dataset observations are sequentially ordered, we examine the subsequent states following each initial state. 
When dealing with unordered expert state transitions, the preferred approach involves establishing temporal order by constructing state transition chains through systematic comparison of each transition's terminal state with the initial states of all remaining transitions.
Since these expert transitions exhibit random distribution and sparsity, complete reconstruction of extended trajectories cannot be guaranteed. 
Nevertheless, we work with the available data to extract meaningful trajectory segments where possible
This trajectory reconstruction process is performed once during the dataset import phase.

For the agent, the situation is considerably different. 
Since we do not possess an imported dataset representing the agent's capabilities, we must investigate possible trajectories within the replay memory. 
We recall that the replay memory is a buffer storing sequential experiences encountered by the agent. 
This makes it ideally suited for traversing through sequential states, with the fundamental assumption that these trajectories inherently reflect the agent's action capabilities.

\begin{figure}
  \centering
\includegraphics[width=0.5\textwidth]{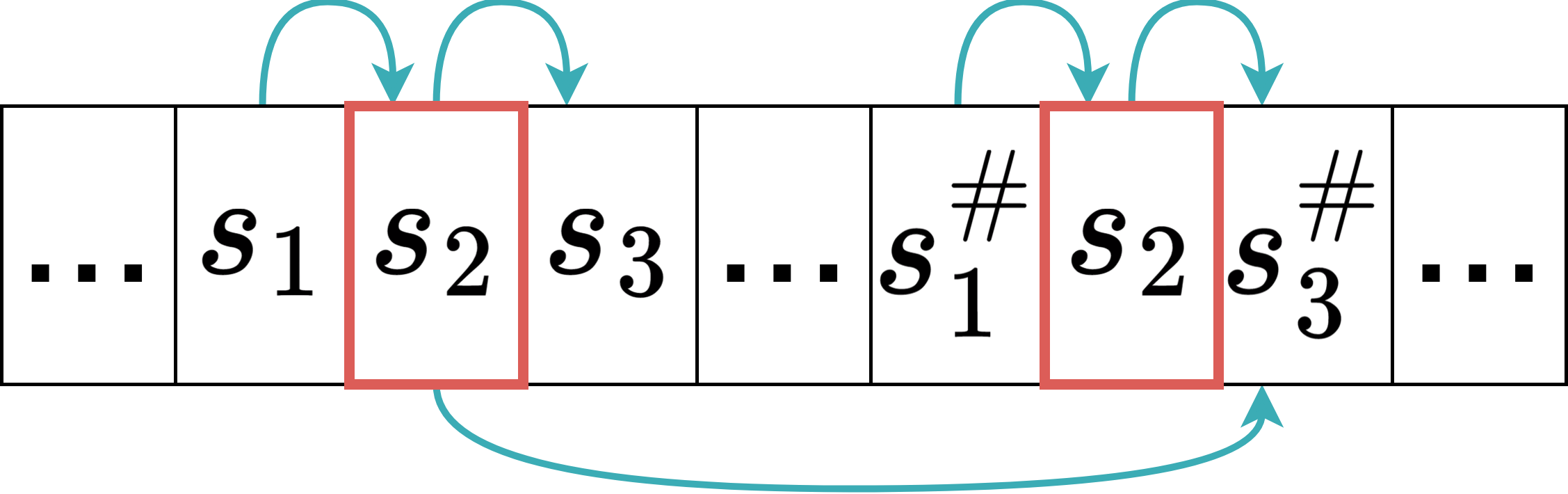}
\caption{Cross-referencing mechanism for trajectory expansion. 
Within a single dataset, similar states occurring at different positions enable trajectory recombination. \textbf{Original path:} Following the natural sequential order (e.g., $s_1 \rightarrow s_2 \rightarrow s_3$ or $s_1^{\#} \rightarrow s_2 \rightarrow s_3^{\#}$). \textbf{New discovered path:} By switching at similar states, alternative trajectories are created by combining segments from different parts of the dataset ($s_1 \rightarrow s_2 \rightarrow s_3^{\#}$). This cross-referencing mechanism expands the set of available paths beyond the original sequences, providing additional options for bridge discovery in heterogeneous action settings.}
\label{fig:cross_reference}
\end{figure}

An additional strategy for populating the trajectory sets involves cross-referencing states across their respective datasets, whether in the replay memory or the expert dataset (see Figure~\ref{fig:cross_reference}). 
This cross-referencing mechanism exploits the property that similar states in both datasets can serve as interchangeable waypoints, enabling trajectory recombination. 
When $s_i \approx s_j$, the behavioural sequence following $s_j$ becomes a viable alternative for the behavioural sequence following $s_i$, effectively expanding the available trajectories beyond the original sequential constraints.


Simply defining how to identify alternative trajectories is insufficient, as we cannot search for alternative paths indefinitely while exhausting computational resources. 
To maintain the search within a controlled computational framework, we complete the definition of infeasibility bridging by introducing the \textit{deep k-n step repair}, a function that traverses bridges with maximum depth of $k$ steps for the agent and $n$ steps for the expert. This approach allows independent control of the maximum search depth for each entity, providing flexibility in balancing computational efficiency with search comprehensiveness.
During such a search, the objective is to identify a common future state shared by both the expert and agent trajectories. 
When this convergence occurs, we establish that a bridge exists which the agent can follow to intercept the expert's behavioural trajectory and maintain alignment with their guidance. 
It should be noted that the maximum depth parameters can be dynamically adjusted during training, either by decreasing their values to avoid unnecessary computational overhead when bridges are consistently discovered through relatively shorter paths, or by increasing their values when bridges cannot be reliably identified with the current search depth.

When a bridge is identified, three values are essential for the learning process. 
For each respective expert transition, we maintain these additional values: the initial action of the bridge $a_{feas}$, the subsequent state of the bridge $s_{feas}$, and the bridge length $l_{feas}$. 
We preserve the bridge length to evaluate which bridges are worth retaining based on their efficiency. 
These values are updated only when a shorter bridge is discovered, as the objective is to identify the minimal path for aligning with the expert's guidance.

It is important to note that this procedure can effectively handle expert datasets containing significant gaps between recorded observations. 
To illustrate this capability, consider a maze navigation scenario where instead of providing the complete trajectory from start to finish, the expert dataset contains only sparse checkpoint states that the expert visited. 
These checkpoints may be considerably distant from one another, yet still provide valuable navigational guidance. 
The bridging mechanism fundamentally addresses these gaps by leveraging the agent's exploration experience to discover feasible paths connecting these sparse expert waypoints. 
Although this requires more extensive exploration, resulting in slower training convergence, the heterogeneous nature of our algorithm enables effective utilization of such sparse expert guidance to reconstruct meaningful learning trajectories.


Algorithmically, to mitigate the computational burden of the complete infeasibility procedure, we recommend conducting infeasibility searches and updates at predetermined intervals rather than at every training step. 
Depending on the sizes of the replay memory and expert dataset, it is advisable to perform these updates at substantial intervals, comparable to those used for target network updates, or even larger intervals when the agent employs conservative exploration strategies. 
Conservative exploration results in reduces the probability of finding new bridges since the last update, making less frequent evaluations more practical.

\subsection{Learning with Infeasibility}

After establishing a method to augment each infeasible expert transition with additional information, we can proceed to explain how the model performs learning with this data. 
When the selected transition is indeed infeasible, we follow the loss function computation of DIIQN as introduced in Section~\ref{sec:diiqn-loss} with two modifications.

The first modification involves a different update for the expert component of the loss function:
\begin{equation}\label{augm_infeas}
    L(\theta_t)_e = \mathbb{E}_{s,a,r,s'}\left[(r + \gamma Q(s_{feas}, \operatorname*{argmax}_{a'} Q(s_{feas}, a'; \theta_t^-); \theta^-_t) - Q(s_e, a_{feas}; \theta_t))^2\right]
\end{equation}
Since the agent cannot execute the infeasible transition presented by the expert, we substitute it with the initial feasible transition from the discovered bridge---the values stored alongside the expert transition during the bridging procedure. 
This approach does not incorporate the complete bridge trajectory into the learning process; rather, we learn only from the first feasible transition. 
This design preserves both the off-policy property of DQN and the single-step decomposition characteristic of temporal difference learning. 
This single step toward the intuitively beneficial direction that leads to expert guidance is expected to provide an effective learning alternative, analogous to how the expert dataset guides the policy toward high-value states.

The second modification pertains to the confidence mechanism. 
Initially, the Q-value divergence component must be adjusted to reflect the feasible action rather than the original expert action:
\begin{equation}
\Delta Q(s_e,a_{feas},a_a) = \sigma\left(\left[Q(s_e, a_{feas}) - Q(s_e, a_a)\right] \cdot \beta\right)
\end{equation}
Additionally, since the newly discovered feasible transition used in place of the expert transition originates from the agent's own experience, we have complete certainty that this exact transition occurred at some point during training. This means that when utilizing the transition $\langle s_e, a_{feas}, s_{feas} \rangle$, there is no uncertainty in the action estimation, as the action was directly observed by the agent.
Consequently, since the action inference error value is always zero, we can eliminate the action inference confidence component $\epsilon(s_e)$ entirely, resulting in the following simplified confidence mechanism for HA-DIIQN:
\begin{equation}
   \Phi(s_e,a_{feas},a_a) = \Delta Q(s_e,a_{feas},a_a) \cdot w(s_e)
\end{equation}
The remaining aspects of the confidence mechanism are applied identically to the standard DIIQN approach:
\begin{equation}
    L(\theta_t) = \Phi(s_e,a_{feas},a_a) L(\theta_t)_e + \left(1-\Phi(s_e,a_{feas},a_a)\right) L(\theta_t)_a
\end{equation}



\section{Experimental Evaluation}
\label{sec:experiments}

In this section, we empirically validate the effectiveness of our proposed methodologies through comprehensive experimentation and analysis. 
We begin by presenting the experimental environments and associated configuration details under which we conducted our evaluation. 
Afterwards, we provide comprehensive analysis of the performance of both proposed approaches, DIIQN and HA-DIIQN, while comparing the baseline version (DIIQN) against other relevant algorithms that address implicit imitation scenarios. 
Finally, we present a parameter sensitivity study that examines how various hyperparameters influence agent performance and establishes recommended ranges for optimal implementation.

\subsection{Experimental Setup}
\label{sec:exper_setup}
For our experimental evaluation, we employ a diverse collection of environments while also introducing environments specifically designed for heterogeneous action settings, as no existing benchmarks adequately address this aspect of the research domain.

\subsubsection{Environments for DIIQN} 
\label{sec:env_diiqn}

\begin{figure}
    \centering
    
    \begin{subfigure}[b]{0.3\textwidth}
        \centering
        \includegraphics[width=\textwidth]{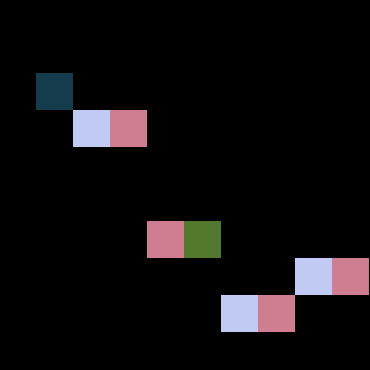}
        \caption{Asterix-v1}
        \label{fig:fig1}
    \end{subfigure}
    \hfill
    \begin{subfigure}[b]{0.3\textwidth}
        \centering
        \includegraphics[width=\textwidth]{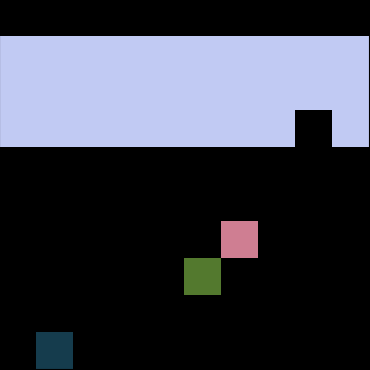}
        \caption{Breakout-v1}
        \label{fig:fig2}
    \end{subfigure}
    \hfill
    \begin{subfigure}[b]{0.3\textwidth}
        \centering
        \includegraphics[width=\textwidth]{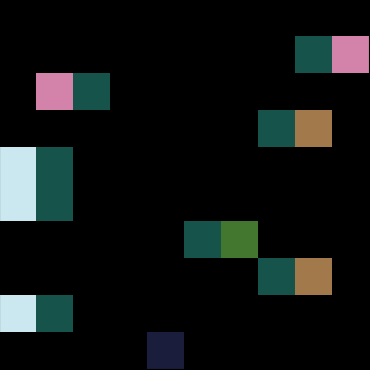}
        \caption{Freeway-v1}
        \label{fig:fig3}
    \end{subfigure}
    
    \vspace{0.5cm}
    
    \hspace{0.148\textwidth}
    \begin{subfigure}[b]{0.3\textwidth}
        \centering
        \includegraphics[width=\textwidth]{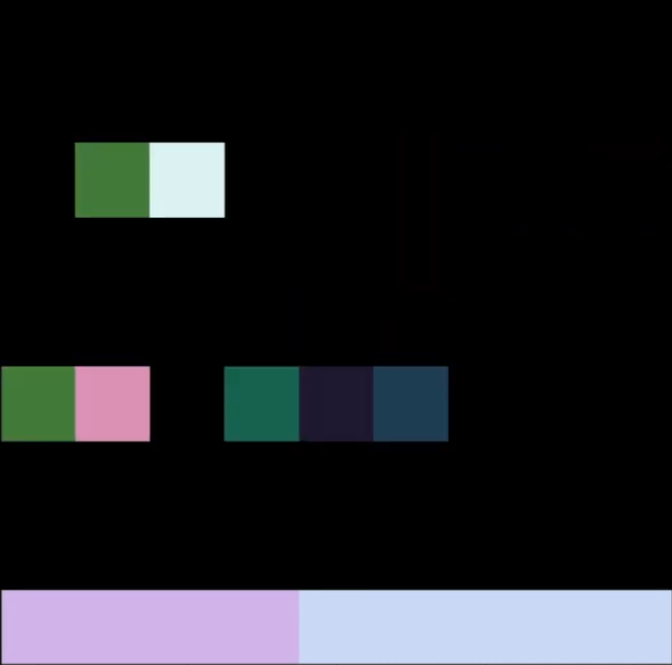}
        \caption{Seaquest-v1}
        \label{fig:fig4}
    \end{subfigure}
    \hspace{-0.013\textwidth}
    \hfill
    \begin{subfigure}[b]{0.3\textwidth}
        \centering
        \includegraphics[width=\textwidth]{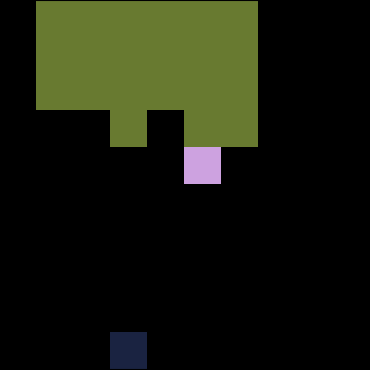}
        \caption{SpaceInvaders-v1}
        \label{fig:fig5}
    \end{subfigure}
    \hspace{0.148\textwidth}
    
    \caption{Visual representation of the MinAtar environments~\cite{young19minatar}, inspired by the original Atari games and their popularity in the DRL domain~\cite{DQN}.}
    \label{fig:minatar}
\end{figure}

Beginning with the environments used to evaluate the standard DIIQN algorithm under homogeneous action settings, we utilize the MinAtar suite~\cite{young19minatar}. 
The MinAtar library provides a collection of environments inspired by classic Atari games, which serve as a standard testbed throughout DRL research~\cite{DQN,kaiser2019model,hosu2016playing}. 
This library creates lightweight representations of five original games while preserving computational complexity through pixel-based state representations. 
This design enables experiments to be conducted with reduced computational requirements while providing meaningful results for algorithm evaluation, as these environments present similar challenges to their original counterparts.

The five MinAtar environments comprise the well-established  Atari games \textit{Asterix}, \textit{Breakout}, \textit{Freeway}, \textit{Seaquest}, and \textit{Space Invaders} (see Figure~\ref{fig:minatar} for visual representation).
Additionally, \textit{sticky-actions} are incorporated, where each agent action may be overridden by the previously executed action, eliminating deterministic behaviour and implicitly introducing planning challenges for the agent.

This environment suite provides an ideal testbed for our evaluation due to several key characteristics: they present reasonable exploration challenges (particularly Seaquest), require meaningful state representation learning, include delayed rewards that challenge the action inference procedure, test algorithm robustness across diverse reward structures and action spaces, and maintain interpretable behaviour patterns suitable for human analysis. 
These environments effectively demonstrate DIIQN's core contributions while ensuring computational tractability for comprehensive experimental analysis. 
Detailed information regarding the state space, action space, and reward functions of each environment can be found in the original publication~\cite{young19minatar}.

\subsubsection{Environments for HA-DIIQN}
\label{sec:env_ha}
For heterogeneous action settings, we cannot employ the same environment set as it is not designed to accommodate heterogeneous action configurations. 
All actions within these environments are essential for successful completion, precluding partial or complete replacement to create alternative action sets. 
Consequently, we introduce two specialized environments: \textit{2D Maze} and \textit{Point Maze}.

\begin{figure}
\centering
    \begin{minipage}{0.26\linewidth}
    \begin{subfigure}{\textwidth}
    \centering
    \includegraphics[width=0.6\linewidth]{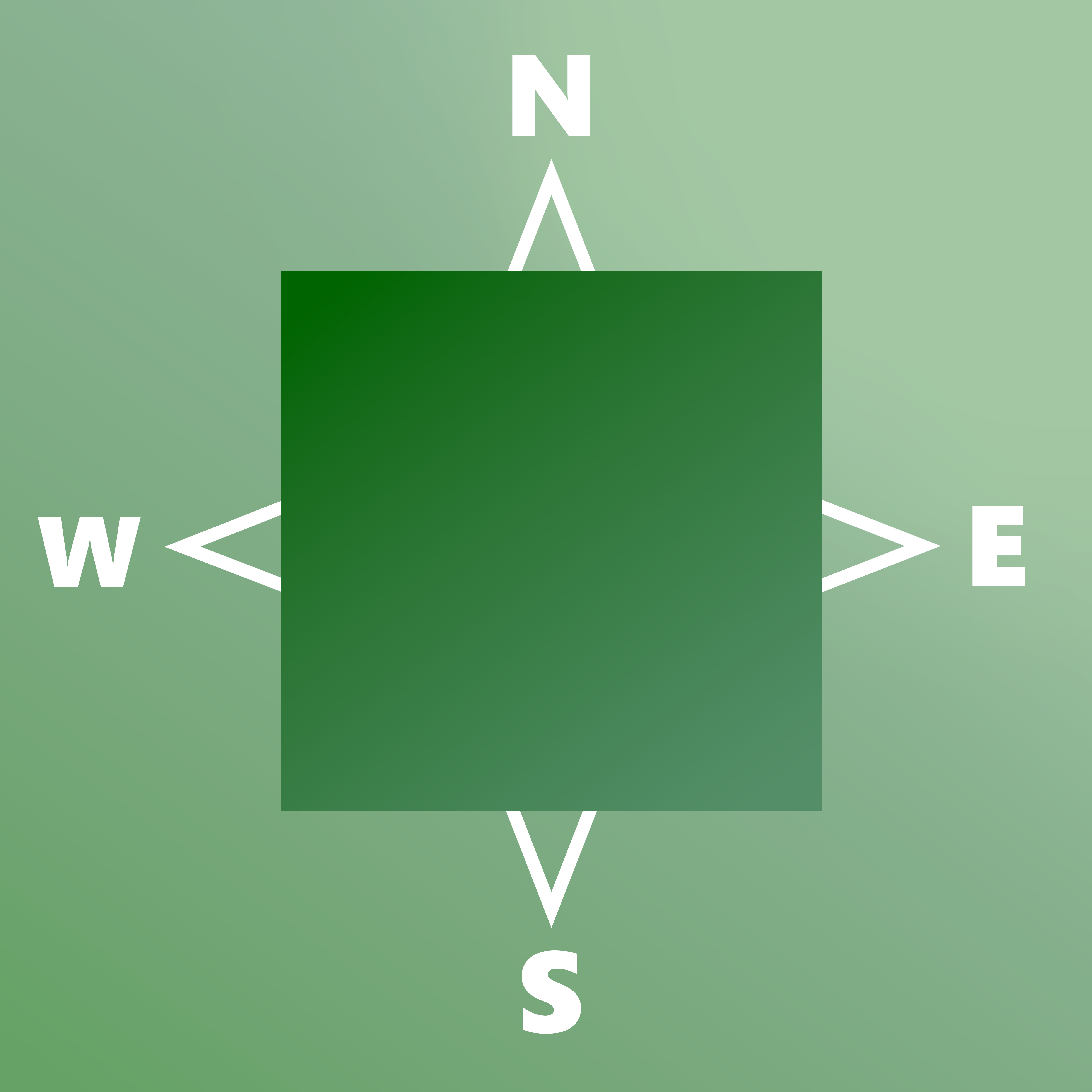}
    \caption{Standard action pattern}
    \label{fig:maze.a}
    \end{subfigure}
    \end{minipage}
    \begin{minipage}{0.26\linewidth}
    \begin{subfigure}{\textwidth}
    \centering
    \includegraphics[width=0.6\linewidth]{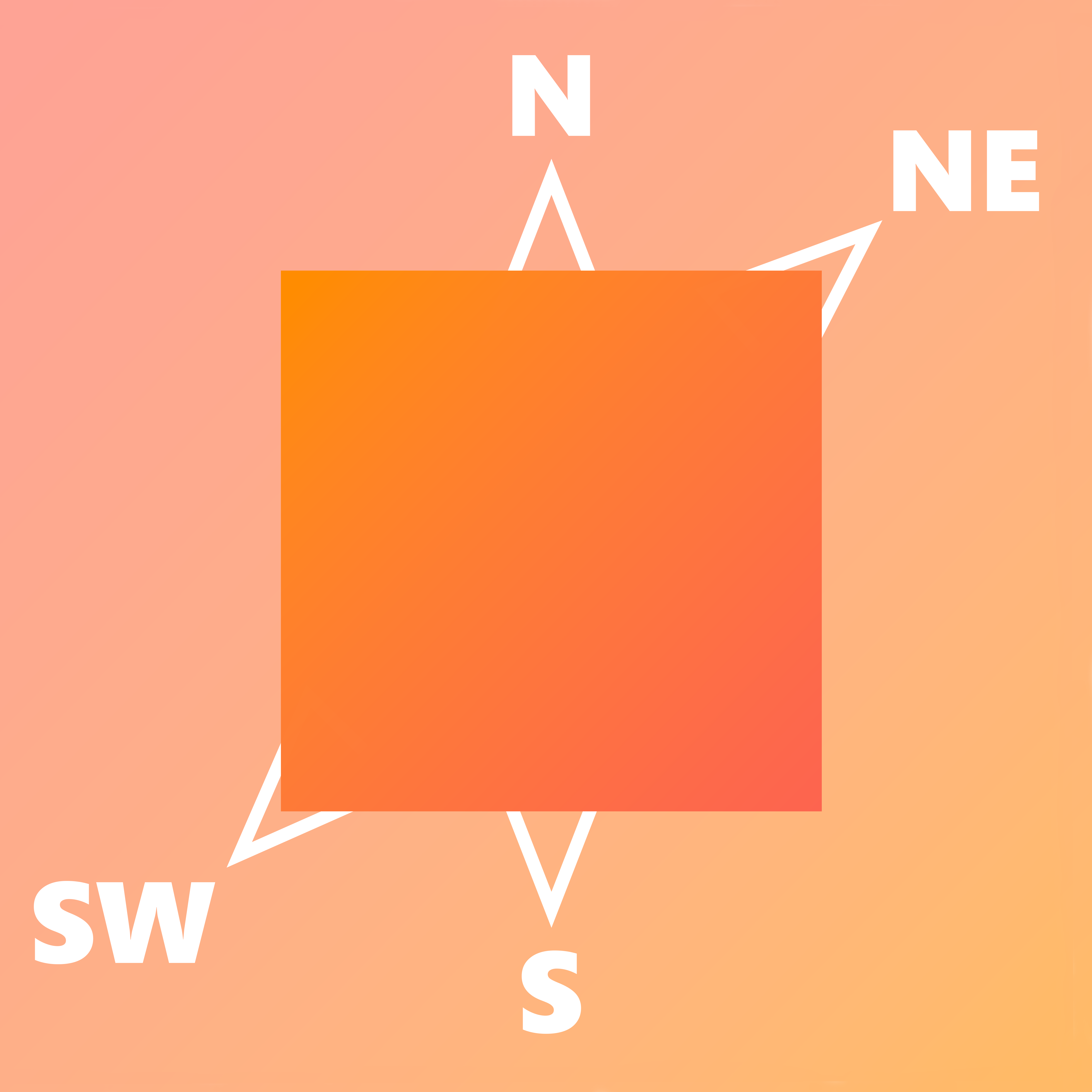}
    \caption{Modified action pattern}
    \label{fig:maze.b}
    \end{subfigure}
    \end{minipage}
    \hfil
    \begin{minipage}{\linewidth}
    \begin{subfigure}{\textwidth}
    \centering
    \includegraphics[width=0.45\linewidth]{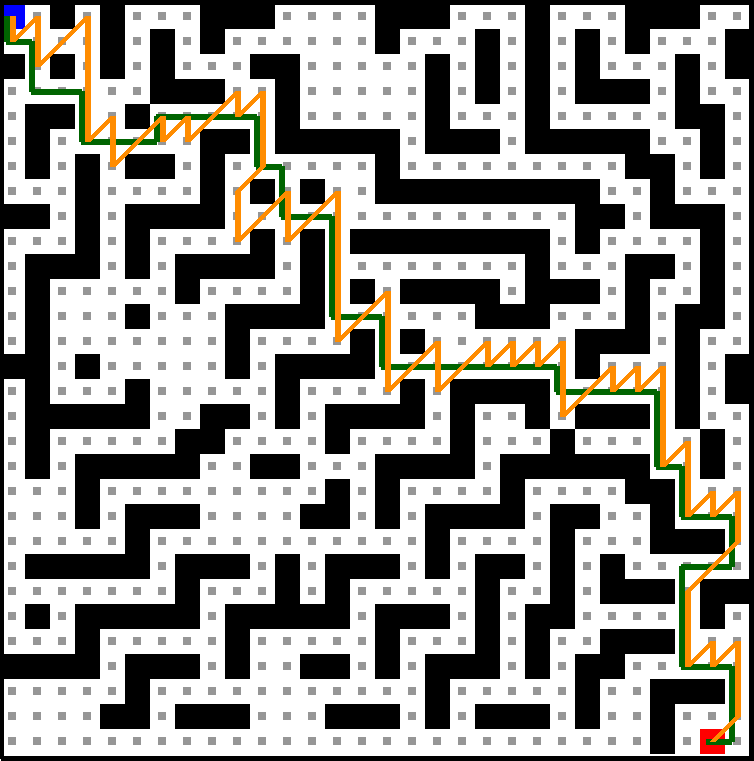}
    \caption{ Visualization of 2D Maze $30\times30$ displaying optimal trajectories for standard action pattern (green) and modified action pattern (orange).}
    \label{fig:maze.c}
    \end{subfigure}
    \end{minipage}
\caption{2D Maze}
\end{figure}

The 2D Maze environment presents an intuitive setting featuring a maze representation with corridors leading to various branching paths, including dead ends, loops, and possibly multiple viable routes to the goal (Figure~\ref{fig:maze.c}). 
The environment contains a designated starting position where the agent spawns and a target endpoint that must be reached. 
An array of walls constrains movement and compels the agent to explore systematically to locate the exit. 
Each time step incurs a negative reward to discourage suboptimal path lengths, while a substantial positive reward is provided upon goal achievement.

Two distinct action sets implement the heterogeneous action space paradigm: the standard set employed by the expert (Figure~\ref{fig:maze.a}) and the modified set used by the agent (Figure~\ref{fig:maze.b}). 
These sets demonstrate partial overlap, illustrating algorithm performance when certain expert guidance remains replicable while other actions require infeasibility procedures to handle inaccessible expert behaviours.

\begin{figure}
\centering
    \begin{minipage}{0.4\linewidth}
    \begin{subfigure}{\textwidth}
    \centering
    \includegraphics[width=0.6\linewidth]{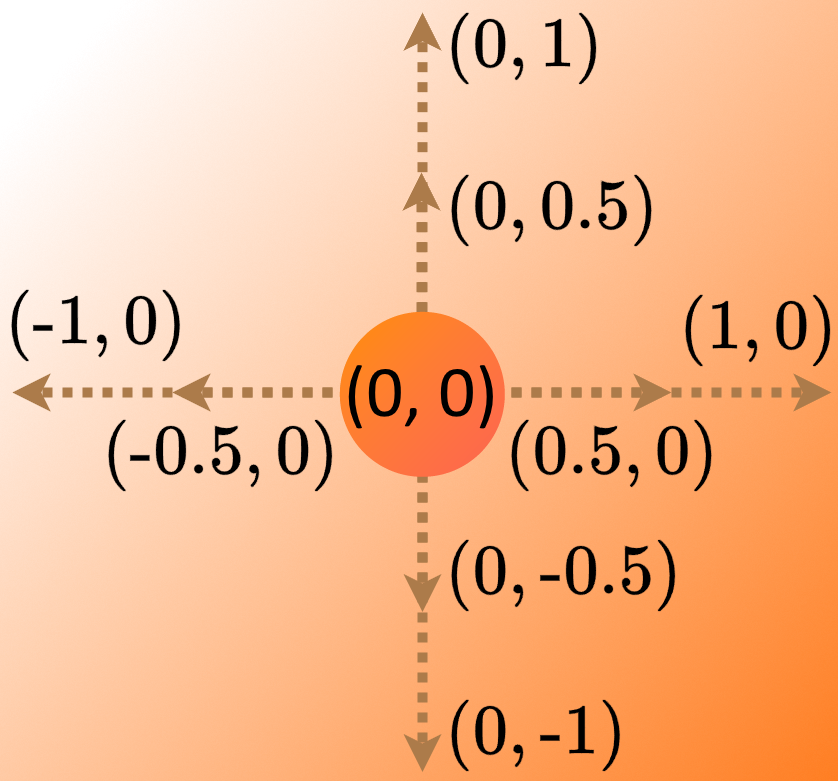}
    \caption{Standard action pattern}
    \label{fig:pointmaze.a}
    \end{subfigure}
    \end{minipage}
    \begin{minipage}{0.4\linewidth}
    \begin{subfigure}{\textwidth}
    \centering
    \includegraphics[width=0.6\linewidth]{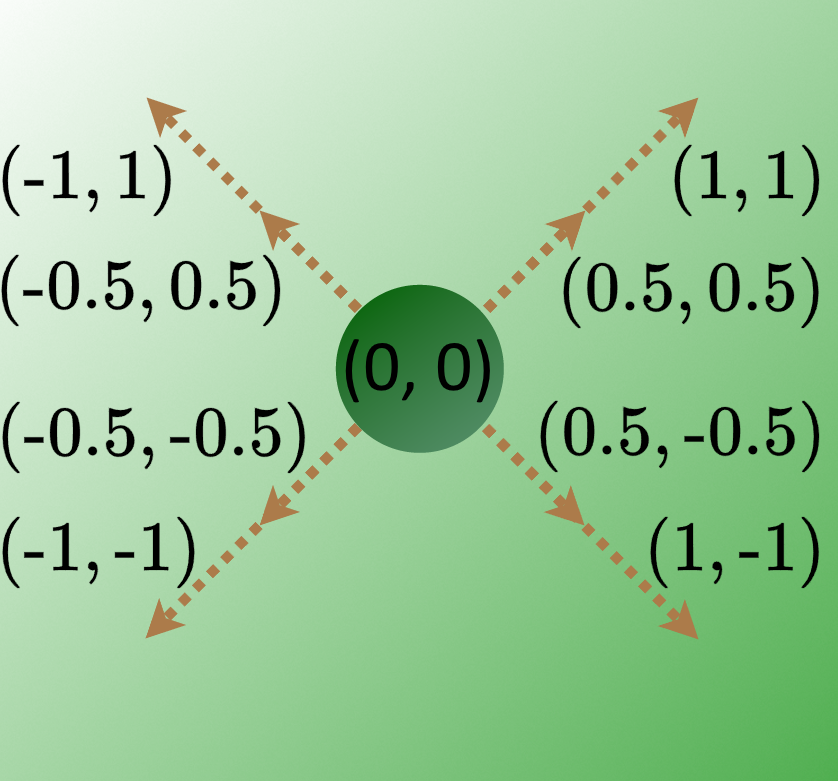}
    \caption{Modified action pattern}
    \label{fig:pointmaze.b}
    \end{subfigure}
    \end{minipage}
    \hfil
    \begin{minipage}{\linewidth}
    \begin{subfigure}{\textwidth}
    \centering
    \includegraphics[width=0.65\linewidth]{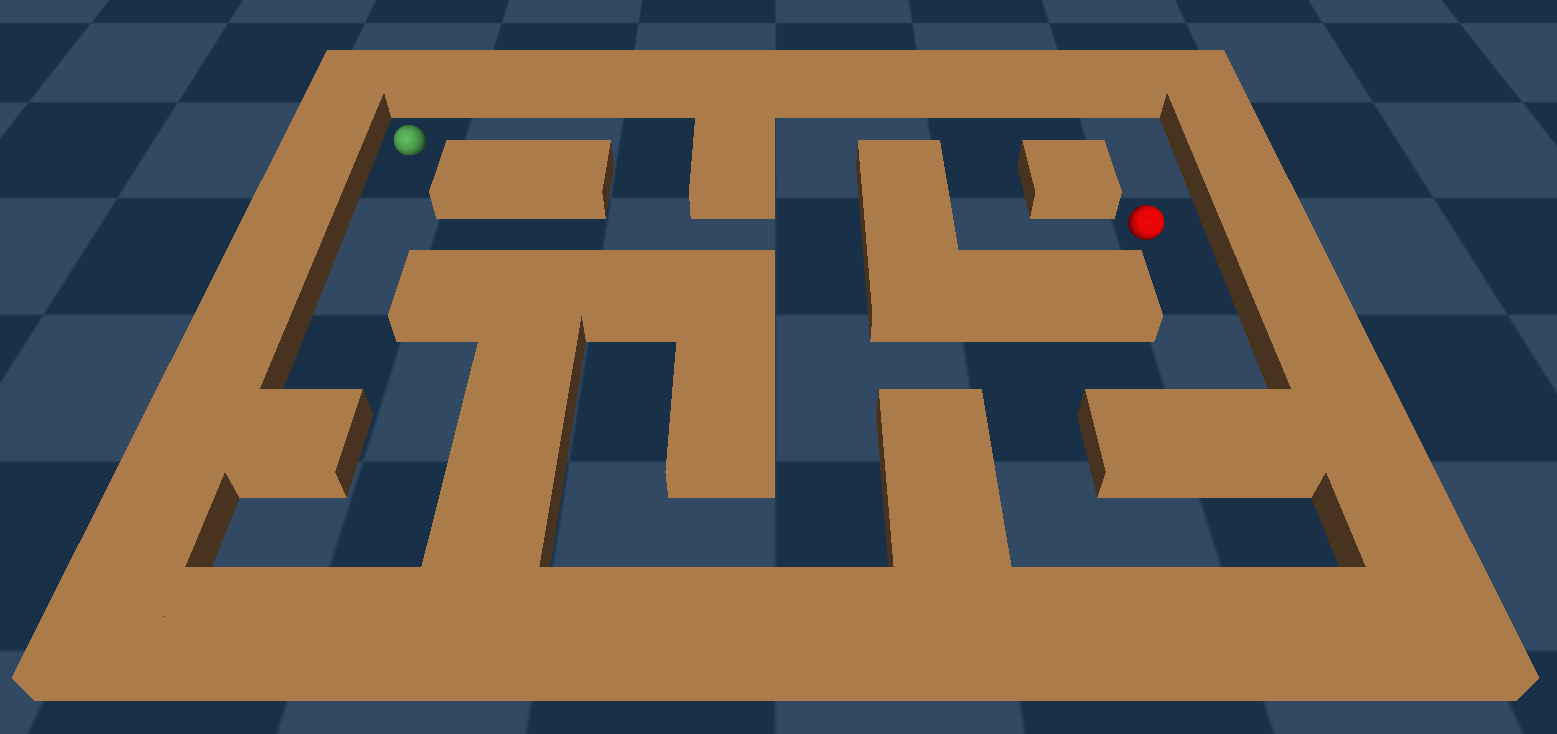}
    \caption{Point Maze navigation environment with agent placement (green) and goal location (red).}
    \label{fig:pointmaze.c}
    \end{subfigure}
    \end{minipage}
\caption{Point Maze}
\end{figure}

We additionally evaluate HA-DIIQN in a more complex environment using D4RL's Point Maze~\cite{fu2020d4rl}, a MuJoCo-based environment~\cite{6386109} featuring a three-dimensional maze with a spherical agent and designated goal location (Figure~\ref{fig:pointmaze.c}). 
The agent controls the sphere through force actuators, navigating to the goal via the most efficient path while minimizing step count. 
The reward structure mirrors the 2D Maze design, penalizing extended trajectories through negative step rewards while providing substantial completion rewards.

As a force-driven MuJoCo environment, it introduces physics-based challenges involving inertia that demand precise control for successful navigation. 
This creates a demanding training environment for the agent that provides clear visual feedback during evaluation. 
The physical dynamics add computational complexity that tests the robustness of our approach under realistic control scenarios.

For the heterogeneous action configurations, we provide two action spaces: the \textit{orthogonal} action space employed by the expert (Figure~\ref{fig:pointmaze.a}) and the \textit{diagonal} action space used by the agent (Figure~\ref{fig:pointmaze.b}). 
Excluding the no-operation action, these spaces share no common actions, demonstrating HA-DIIQN's capabilities when confronted with completely disjoint expert datasets that would provide no value under conventional imitation learning approaches.

\subsubsection{Suboptimal Expert Dataset Collection}
To demonstrate our algorithm's ability to leverage diverse behavioural patterns, we collect expert data from multiple agents exhibiting different policies. 
Each agent is trained using DQN until reaching our target suboptimal episodic reward during evaluation, at which point we record two complete episodes. 
These individual datasets are combined to form our complete expert dataset; details regarding final dataset sizes and their impact on performance are analyzed in Section~\ref{sec:parameter_eval}.

As previously discussed, the target episodic reward represents suboptimal performance, deliberately chosen to fall short of the final converged performance achievable through extended DQN training. 
This suboptimality requirement introduces an interesting consequence: increased behavioural diversity among experts.
While optimal policies in most environments exhibit limited variation (as there are few distinct ways to solve a problem optimally), suboptimal policies demonstrate considerably greater diversity. 
Multiple different approaches can achieve mediocre performance, resulting in a naturally diverse expert dataset. 
This diversity would typically confuse traditional imitation learning approaches that lack the capacity to critically evaluate expert guidance. 
However, our integration of DRL enables the agent to filter suboptimal behaviours and develop superior policies despite—and indeed, benefiting from—this varied guidance.

\subsection{Experimental Results}

We proceed to evaluate the performance of our proposed framework, DIIQN and HA-DIIQN, comparing them against baseline methods and competing algorithms.

\subsubsection{DIIQN}

\begin{figure}
    \centering
    
    \begin{subfigure}[b]{0.46\textwidth}
        \centering
        \includegraphics[width=\textwidth]{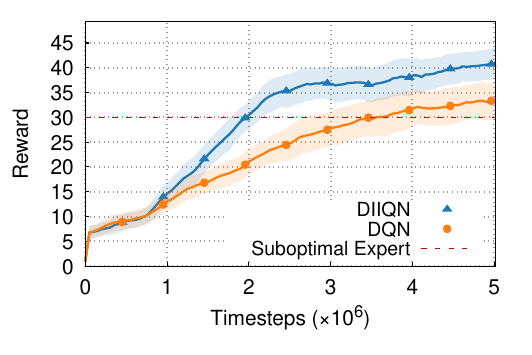}
        \caption{Asterix-v1}
        \label{fig:fig1_diiqn}
    \end{subfigure}
    \hfill
    \begin{subfigure}[b]{0.46\textwidth}
        \centering
        \includegraphics[width=\textwidth]{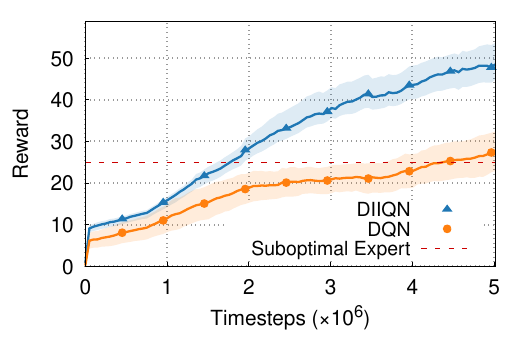}
        \caption{Breakout-v1}
        \label{fig:fig2_diiqn}
    \end{subfigure}
    
    \vspace{0.3cm}
    
    \begin{subfigure}[b]{0.46\textwidth}
        \centering
        \includegraphics[width=\textwidth]{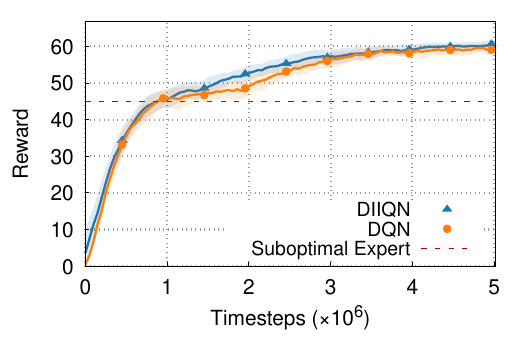}
        \caption{Freeway-v1}
        \label{fig:fig3_diiqn}
    \end{subfigure}
    \hfill
    \begin{subfigure}[b]{0.46\textwidth}
        \centering
        \includegraphics[width=\textwidth]{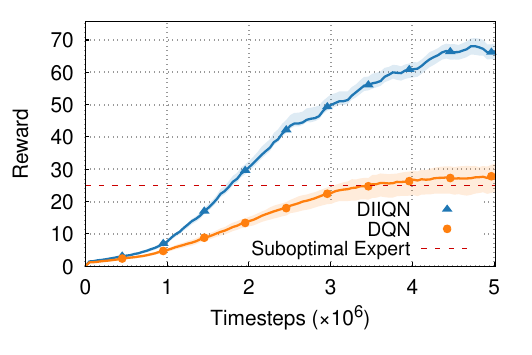}
        \caption{Seaquest-v1}
        \label{fig:fig4_diiqn}
    \end{subfigure}
    
    \vspace{0.3cm}
    
    \begin{subfigure}[b]{0.46\textwidth}
        \centering
        \includegraphics[width=\textwidth]{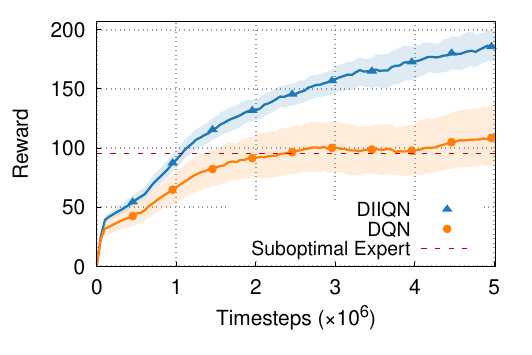}
        \caption{SpaceInvaders-v1}
        \label{fig:fig5_diiqn}
    \end{subfigure}
    
    \caption{Average episodic reward on MinAtar games over 10 runs, comparing DIIQN and DQN. Shaded regions indicate standard deviation across runs. Each point corresponds to the total reward obtained in an episode at that training step. The horizontal dashed lines represents the performance of the suboptimal expert policy from which observations were collected.}
    \label{fig:all_figures_diiqn}
\end{figure}

The DIIQN paradigm fundamentally combines DRL with IL to leverage the complementary strengths of both approaches: accelerated training through imitation learning and pursuit of optimality through reinforcement learning. 
In this section, we demonstrate the performance of our algorithm compared against both pure DRL and IL techniques, in the MinAtar environments as introduced in Section~\ref{sec:env_diiqn}.

\paragraph{Competing against a Deep Reinforcement Learning baseline}

We now present the performance of DIIQN when put against DQN.
We naturally employ DQN as a DRL baseline, given that DIIQN builds upon its foundational architecture. 
DQN serves as an ideal comparison point due to its widespread adoption and recognition within the community, providing clear interpretability of performance differences. 
Furthermore, comparing against DQN ensures fairness in evaluation, as both algorithms can be configured with identical hyperparameters for shared components, isolating the contribution of our proposed extensions without ambiguous factors related to parameter tuning.

\begin{figure}
    \centering
    
    \begin{subfigure}[b]{0.48\textwidth}
        \centering
        \includegraphics[width=\textwidth]{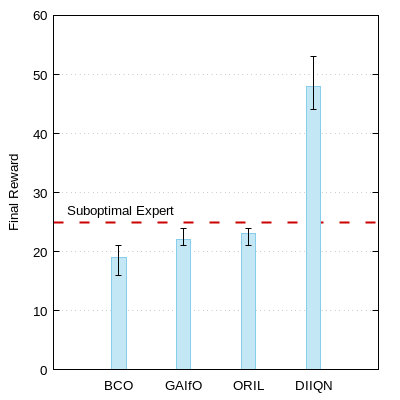}
        \caption{Breakout-v1}
        \label{fig:fig1_comp}
    \end{subfigure}
    \hfill
    \begin{subfigure}[b]{0.48\textwidth}
        \centering
        \includegraphics[width=\textwidth]{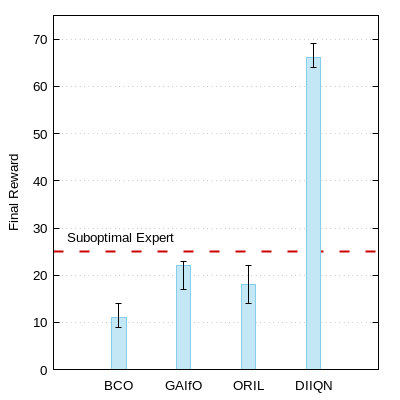}
        \caption{Seaquest-v1}
        \label{fig:fig2_comp}
    \end{subfigure}
    \begin{subfigure}[b]{0.48\textwidth}
        \centering
        \includegraphics[width=\textwidth]{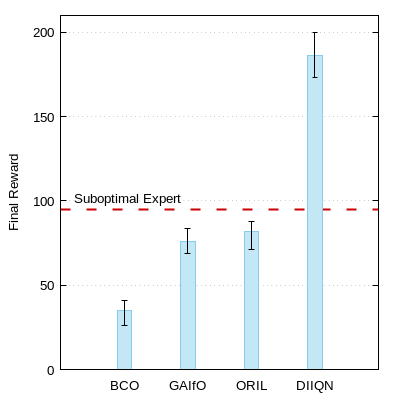}
        \caption{SpaceInvaders-v1}
        \label{fig:fig3_comp}
    \end{subfigure}
    
    \caption{Final episodic reward on MinAtar games, averaged over 10 runs, comparing DIIQN with BCO, GAIfO, and ORIL. Error bars indicate standard deviation across runs. The horizontal dashed line represents the performance of the suboptimal expert policy from which observations were collected.}
    \label{fig:all_figures_comp}
\end{figure}

Our results are presented in Figure~\ref{fig:all_figures_diiqn}.
The evaluation curves of DIIQN and DQN are 
over a training horizon of $5$M steps. 
Under identical parameterization for shared hyperparameters, the DIIQN agent leverages suboptimal expert guidance to substantially improve learning efficiency and final performance. 
Excluding Freeway, DIIQN outperforms DQN across all remaining environments, achieving $24$-$130\%$ higher episodic returns. 
By exploiting expert experience, even from suboptimal demonstrators, superior reward acquisition is achieved within the same training duration, demonstrating the effectiveness of implicit imitation integration.

The exception observed in Freeway stems from its unique environmental characteristics. 
Freeway employs a fixed episode length rather than the infinite horizon structure present in other environments, making optimal behaviour relatively straightforward to discover. 
The absence of infinite-horizon challenges that would otherwise amplify minor policy flaws and terminate episodes prematurely, reduces the complexity of achieving near-optimal performance. 
Therefore, the edge provided by implicit imitation learning\footnote{Had the approach been an {\em explicit} imitation learning one---i.e., if the agent had access to the expert actions---then we expect that the agent would benefit from imitating the expert even in this easy-to-learn game setting.} is simply not required in this environment.
A simple DRL approach is enough to discover the optimal policy. Indeed, DIIQN reduces to behaving such as standard DQN (its confidence mechanism decides that expert guidance is not required), and thus
its learning curve closely resembles that of standard DQN in this game.
Further analysis of the confidence mechanism behaviour is provided in Section~\ref{sec:parameter_eval}.

\paragraph{Competing against Imitation Learning} 
To evaluate DIIQN against other imitation learning approaches, we compare final performance with three established methods: Behavioural Cloning from Observation (BCO)~\cite{ijcai2018p687}, Generative Adversarial Imitation from Observation (GAIfO)~\cite{torabi2018generative}, and Offline Reinforced Imitation Learning (ORIL)~\cite{zolna2020offline}. 
All methods operate within the implicit imitation learning paradigm, utilizing only state observations as expert samples without action labels.
BCO represents a foundational approach included to demonstrate baseline performance using supervised learning principles for action inference. 
GAIfO exemplifies high-performing methods that combine online environment interaction with adversarial learning, building upon the well-established GAIL framework. 
ORIL employs a sophisticated approach that trains a reward function by contrasting expert data against unlabeled trajectories of unknown origin, subsequently using these learned reward signals for agent training.

    
    

The final episodic rewards for all algorithms are presented in Figure~\ref{fig:all_figures_comp}. We conduct experiments across three environments: Breakout, Seaquest, and Space Invaders. 
DIIQN demonstrates substantial performance advantages over all competing methods. 
This performance gap primarily stems from the suboptimal nature of the expert dataset employed in our evaluation. 
Since imitation learning techniques fundamentally aim to replicate expert performance, they naturally inherit the limitations and suboptimalities present in the demonstration data.
It is clear that the expert dataset size proves sufficient for GAIfO and ORIL to approximate expert-level performance, even though neither method can surpass the expert's capabilities. 
On the other hand, DIIQN {\em outperforms the expert} by a large margin, by leveraging its own exploratory experiences that enable the agent to discover superior policies that surpass the quality of the provided observations. 
This capability to improve beyond expert performance constitutes a fundamental advantage of our approach.

\subsubsection{HA-DIIQN}
Proceeding to the heterogeneous extension of DIIQN, we evaluate HA-DIIQN's performance using the environments with heterogeneous action spaces introduced in Section~\ref{sec:env_ha}.
We compare HA-DIIQN against two baselines, DQN and the standard DIIQN. 
The latter is simply DIIQN, and
thus does 
not account for potentially infeasible expert transitions during training, as HA-DIIQN does.
We do not pit our HA-DIIQN approach against
the previously evaluated BCO, GAIfO, and ORIL,
as these would, naturally, exhibit performance characteristics similar to standard DIIQN, since they  do not 
make any provisions for heterogeneous action spaces either.

\begin{figure}[pos=!htbp]
    \centering
    \begin{subfigure}[b]{0.49\textwidth}
        \centering
        \includegraphics[width=\textwidth]{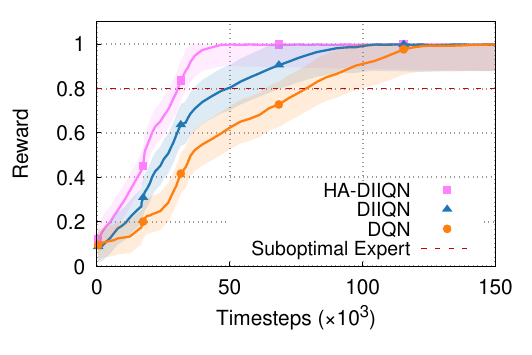}
        \caption{2D Maze}
        \label{fig:fig1_ha}
    \end{subfigure}
    \begin{subfigure}[b]{0.49\textwidth}
        \centering
        \includegraphics[width=\textwidth]{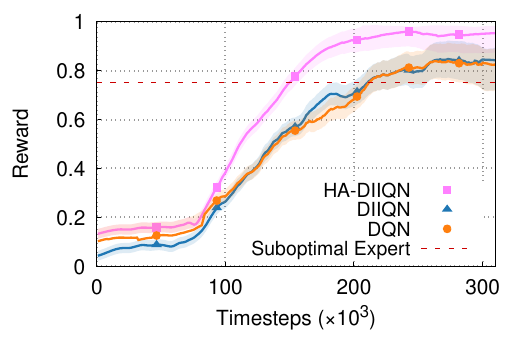}
        \caption{Point Maze}
        \label{fig:fig2_ha}
    \end{subfigure}
    \caption{Average episodic reward on 2D Maze and Point Maze environments over 10 runs, comparing HA-DIIQN, DIIQN and DQN. Shaded regions indicate standard deviation across runs. Each point corresponds to the total normalized reward obtained in an episode at that training step. The horizontal dashed lines represent the performance of the heterogeneous-action suboptimal expert policy from which observations were collected.}
    \label{fig:all_figures_ha}
\end{figure}

Figure~\ref{fig:all_figures_ha} depicts our results in these domains.
Across both environments, 2D Maze (Figure~\ref{fig:fig1_ha}) and Point Maze (Figure~\ref{fig:fig2_ha}), HA-DIIQN demonstrates superior performance compared to both baseline variants, as its bridging mechanism allows it to overcome the infeasibility of certain expert actions, allowing it to learn even from these actions. 
In the 2D Maze, a fully discrete environment where optimal behaviour is achievable, HA-DIIQN learns $52\%$ faster than DIIQN and $64\%$ faster than DQN. 
Specifically HA-DIIQN reaches optimal performance in $43$K timesteps, while DIIQN in $90$K and DQN in $121$K.
DIIQN maintains this modest convergence time performance advantage over DQN due to the partial action space overlap between expert and agent (see Figure~\ref{fig:maze.c}), which, if only partial, allows it to employ its imitation learning capabilities to benefit from expert behaviour when possible. (By contrast, as we noted earlier, HA-DIIQN can benefit from potentially {\em all} expert observations---i.e., including those relating to infeasible actions that can be inferred via its bridging mechanism).

In Point Maze, HA-DIIQN clearly surpasses both variants in terms of final reward. 
HA-DIIQN achieves a final episodic reward of $0.95$ within $211$K training timesteps, while DIIQN and DQN require approximately $300$K timesteps to reach $0.83$ final reward.
In this environment, DIIQN and DQN exhibit nearly identical performance due to complete action space disjunction between agent and expert (as seen in Figure~\ref{fig:pointmaze.c}), rendering the expert dataset effectively unusable for DIIQN. 
The expert transitions cannot be replicated, consistently producing large action error estimations.
However, HA-DIIQN can exploit its bridging mechanism to overcome this difficulty in this environment also.

Both environments were previously evaluated in an earlier version of HA-DIIQN~\cite{Chrysomallis_Chalkiadakis_Papamichail_Papageorgiou_2025} using a simpler algorithmic variant. 
That version employed a naive binary switching mechanism that selected exclusively between expert or agent loss functions, without accounting for the nuanced edge cases handled by the refined confidence mechanism introduced in Section~\ref{sec:diiqn-loss}. 
To validate the improvements provided by the current framework, we reran all experiments with the updated algorithm. 
The results demonstrate substantially improved stability, with significantly reduced standard deviation and more consistent learning curves (cf.~Figures 8 and 9 in~\cite{Chrysomallis_Chalkiadakis_Papamichail_Papageorgiou_2025}).
This validates that the dynamic confidence weighting mechanism contributes meaningfully to both the homogeneous (DIIQN) and heterogeneous (HA-DIIQN) variants of our framework.

\subsection{Parameter Sensitivity Analysis}
\label{sec:parameter_eval}

\begin{figure}
    \centering
    
    \begin{subfigure}[b]{0.48\textwidth}
        \centering
        \includegraphics[width=\textwidth]{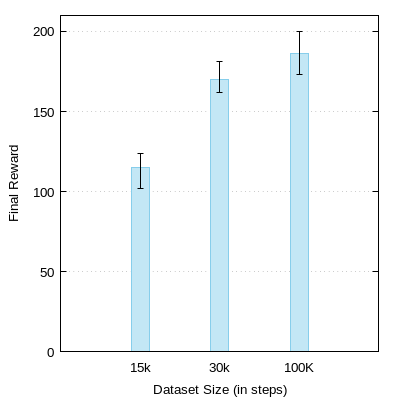}
        \caption{Dataset size}
        \label{fig:fig1_abl}
    \end{subfigure}
    \hfill
    \begin{subfigure}[b]{0.48\textwidth}
        \centering
        \includegraphics[width=\textwidth]{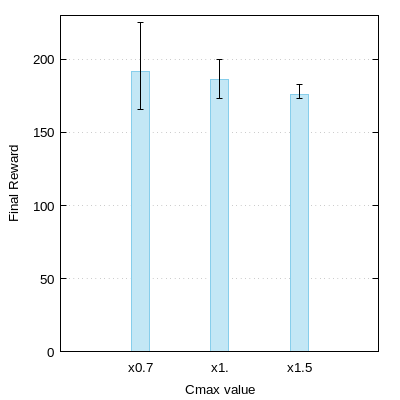}
        \caption{$C_{max}$}
        \label{fig:fig2_abl}
    \end{subfigure}
    \begin{subfigure}[b]{0.48\textwidth}
        \centering
        \includegraphics[width=\textwidth]{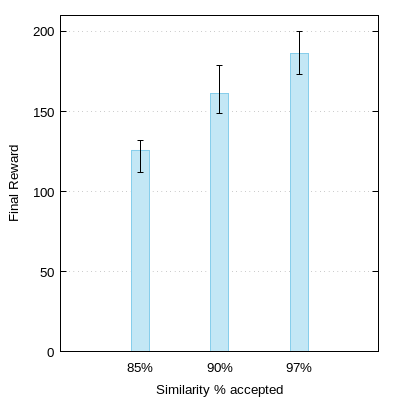}
        \caption{$\tau_{similar}$}
        \label{fig:fig3_abl}
    \end{subfigure}
    \hfill
    \begin{subfigure}[b]{0.48\textwidth}
        \centering
        \includegraphics[width=\textwidth]{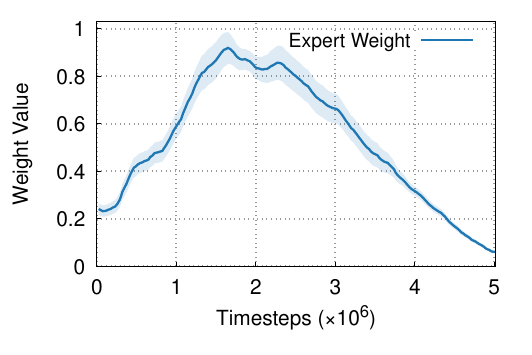}
        \caption{Expert Weight $\Phi$}
        \label{fig:fig4_abl}
    \end{subfigure}
    
    \caption{Final episodic reward of DIIQN under different hyperparameter settings (first three subfigures) and behavioural curve of the expert weight (last subfigure). Results are averaged over 10 runs. Evaluations are conducted on Space Invaders. Error bars indicate standard deviation.}
    \label{fig:all_figures_abl}
\end{figure}

We now present a comprehensive parameter sensitivity analysis that provides deeper insights into various components of our proposed algorithm (Figure~\ref{fig:all_figures_abl}). 
To ensure result stability, we conduct all experiments in this study in a single environment: the MinAtar {\em Space Invaders} game setting. 
This environment was selected due to several advantageous characteristics: its non-restrictive termination condition (ending only upon player failure rather than at a fixed time point), the high skill level required for optimal performance, its dynamic difficulty progression (accelerating gameplay as enemies are eliminated, demanding varying response times), and its capacity for continuous performance improvement through subtle behavioural refinements.
Importantly, this environment exhibits a high skill ceiling that remains unreachable within our training horizon, allowing us to effectively demonstrate performance differences across parameter configurations. 
We note that all experiments utilize guidance from a suboptimal expert achieving an episodic reward of $95$.

We investigate the following parameters: a) expert dataset size, b) the maximum training update threshold $c_{max}$ for expert state regions as introduced in Section~\ref{sec:diiqn-loss}, and c) the similarity percentage threshold $\tau_{similar}$ from Section~\ref{sec:expert_sampling} for expert sample selection during KNN search.
For each parameter, we evaluate three different configurations, presenting both the final reward achieved and corresponding standard deviation. 
Additionally, we provide behavioural analysis of the expert weight $\Phi(s_e,a_{feas},a_a)$ to illustrate how expert influence affects the training procedure and to what degree this influence persists throughout learning.

We begin by examining the expert weight $\Phi$ behaviour throughout training (Figure~\ref{fig:fig4_abl}). 
We recall that $\Phi$ represents the factor determining the relative contribution of expert guidance versus agent experience at each training step. 
Higher values indicate that the model prioritizes learning from expert guidance, while lower values indicate reliance on the agent's own experiences.
Before presenting our parameterization experiments, we illustrate how this loss function weight evolves across the complete training horizon to provide intuitive understanding. 
We observe a consistent steep increase during the initial $1.5$M steps. 
While substantial expert guidance is expected during early training phases, ensuring the quality of this information remains crucial. 
The confidence mechanism introduced in DIIQN addresses this through two key restrictions: the discount weight $w(s_e)$ (Equation~\ref{disc_weight}) ensures accurate Q-value representations in the investigated state space regions, and the inference reliability $\epsilon(s_e)$ (Equation~\ref{inference_rel}) validates the trustworthiness of inferred actions. 
These restrictions delay full expert sample utilization slightly beyond initial expectations, but guarantee data reliability.
Subsequently, during the next $1$M steps, a stable plateau emerges, demonstrating high-capacity exploitation of expert guidance once sufficient confidence has been established. 
This is followed by a gradual decline over the final $2.5$M steps, as the agent develops policies superior to the expert. 
The agent recognizes diminishing returns from expert guidance, resulting in near-zero weights by training completion.
This progression represents ideal DIIQN loss function behaviour and generalizes across environments when appropriately tuned. 
In our experiments, we have identified and recommend achieving a similar curve profile to ensure correct DIIQN implementation and optimal performance.

Expert dataset size, as anticipated, substantially influences the final reward achievable through DIIQN.
In this analysis, we examine the relationship between dataset size and policy improvement, investigating how increasing sample quantities enhances state space coverage, exposes the agent to more diverse scenarios, improves generalization by enabling extraction of underlying behavioural patterns, and reduces variance in learned policies. 
The objective is to identify the threshold beyond which additional expert samples provide minimal marginal benefit.
We experiment with $15$K, $30$K, and $100$K expert samples, where each sample represents a single state transition (Figure~\ref{fig:fig1_abl}). 
Although standard deviation ranges remain comparable across configurations, final rewards exhibit notable differences. 
With $15$K samples, we observe satisfactory performance reaching an episodic reward of $115$, surpassing the suboptimal expert threshold of $95$. 
Doubling the dataset to $30$K samples produces a substantial improvement to $170$, demonstrating that increased guidance enables the agent to bridge larger performance gaps toward superior behaviours.
However, expanding to $100$K expert samples yields only modest additional gains, achieving a final reward of $186$. 
This plateauing effect indicates diminishing returns beyond a certain dataset size threshold. 
For an environment of Space Invaders' complexity, the substantial improvement from $15$K to $30$K samples demonstrates clear value in additional expert guidance. However, increasing from $30$K to $100$K samples (more than tripling the dataset) produces only marginal gains ($16$ additional reward points). 
This suggests that $30$K samples provide sufficient state space coverage and behavioural diversity for effective learning in this environment. 
Beyond this point, additional expert data offers limited new information, as the agent has already been exposed to representative scenarios necessary for policy development.

Proceeding to the $c_{max}$ analysis, we examine different thresholds for the number of training updates required for each expert state region before achieving maximum confidence (Figure~\ref{fig:fig2_abl}). 
This parameter determines how early the model can reliably represent Q-value dynamics and use these representations to balance expert guidance against the agent's self-discovered capabilities. 
We evaluate three configurations: $100$K, $150$K, and $225$K training updates.
These values indicate that when an expert state (or closely similar state) has been selected for network training $c_{max}$ times, we have established maximum confidence in the Q-value estimates for that region.
Our results demonstrate that lower thresholds, such as $100$K, enable earlier expert guidance utilization, producing immediate performance gains that yield marginally higher final rewards. 
However, the standard deviation patterns reveal a critical consideration.
By trusting Q-value estimates at earlier training stages, the agent becomes susceptible to premature value function approximations, resulting to substantially larger standard deviation in final policy performance. 
This produces unstable learning behaviour that we strongly discourage. 
This instability can be mitigated by employing higher $c_{max}$ values, such as $150$K or $225$K, even at the cost of modest reductions in final reward. 
The trade-off favors stability and reliability over marginal performance gains.
It is important to note that excessively high $c_{max}$ values should also be avoided, as they would severely restrict expert influence throughout training. 
The evaluated range represents appropriate choices when considered relative to the overall training horizon of $5$M steps, balancing timely expert integration with sufficient confidence in value function estimates.

We conclude our parameter evaluation by investigating the similarity threshold component $\tau_{similar} $ of the expert sampling procedure (Figure~\ref{fig:fig3_abl}). 
We recall that after performing a KNN search to identify the nearest expert samples to the agent's current state, we obtain a ranked set of candidates based on proximity. 
However, proximity in the distance metric does not guarantee meaningful similarity, since the nearest neighbors may still represent substantially different states from the agent's actual situation. 
To ensure that only genuinely relevant expert samples are utilized for training, we introduced a similarity threshold that defines acceptable similarity bounds.
This similarity threshold can be expressed as a percentile when the state space is normalized (see Appendix~\ref{apdx:distances} for calculation details). 
We evaluate three acceptance thresholds: $85\%$, $90\%$, and $97\%$. 
While the KNN search yields identical candidate sets across configurations, the similarity percentile determines which candidates are retained for training.
Lower thresholds permit more frequent expert-guided training, as relaxed boundaries allow a larger pool of expert samples to qualify. 
However, this increased quantity comes at the cost of reduced quality.
Accepted samples may not closely represent the agent's actual state, leading to suboptimal action inferences and less relevant guidance. 
The experimental results clearly demonstrate this trade-off: increasing the similarity requirement from $85\%$ (achieving a final reward of $126$) to $97\%$ (achieving $186$) produces approximately $50\%$ improvement in final performance.
We note that in our experimental setting, the similarity threshold has substantial influence due to the environment's sensitive state representation. 
In Space Invaders, individual pixels can carry high semantic importance (such as player position), making small state differences meaningful. 
Maintaining high similarity thresholds is recommended when sufficient expert data is available. 
However, when working with limited expert datasets, relaxing the threshold may become necessary to ensure adequate training signal, accepting the quality-quantity trade-off inherent in this decision.

In summary, this parameter sensitivity analysis reveals three key lessons:
\begin{itemize}
    \item \textbf{Expert dataset size:} Beyond $30$K samples, additional data provides diminishing returns. It is recommended to prioritize dataset diversity and coverage over quantity.
    \item \textbf{Confidence threshold $c_{max}$:} Higher values ($150$K-$225$K) are strongly recommended to ensure training stability, even at modest performance costs, as premature confidence leads to high variance and unreliable learning.
    \item \textbf{Similarity threshold $\tau_{similar}$:} High thresholds ($>95\%$) significantly improve performance by ensuring quality expert guidance, though lower values may be necessary when working with limited datasets.
\end{itemize}
Overall, DIIQN demonstrates robustness across diverse parameter configurations, with its adaptive confidence mechanism enabling effective expert leverage while maintaining the capacity to exceed expert performance.


\section{Conclusions and Future Directions}
\label{sec:conclusions}

This work addresses a critical gap in imitation learning: the ability to learn from observation-only expert datasets that may be suboptimal, while simultaneously handling scenarios where expert and agent possess different action capabilities. 
We introduced DIIQN, a deep reinforcement learning framework that integrates implicit imitation learning to leverage expert guidance while maintaining the capacity to surpass expert performance through environmental interaction. 
Our HA-DIIQN extension further enables learning from experts with heterogeneous action sets, a previously unaddressed scenario in implicit imitation learning literature.
The framework's core contributions include: a) an action inference procedure that reconstructs expert actions from state-only observations through online exploration; b)  augmented loss functions that integrate expert guidance into the DQN training paradigm; c)  a dynamic confidence mechanism that adaptively balances expert-guided and self-directed learning based on Q-value divergence, training frequency, and action inference reliability; and d) a bridging mechanism that identifies feasible alternative pathways when expert transitions are infeasible for the agent.

Our experimental evaluation across diverse environments demonstrates substantial performance improvements. 
In homogeneous action settings, DIIQN achieves up to $130\%$ higher episodic returns compared to standard DQN across MinAtar environments, while consistently surpassing other implicit imitation learning methods (BCO, GAIfO, ORIL) that cannot exceed suboptimal expert performance.
In heterogeneous action scenarios, HA-DIIQN successfully bridges action space disparities, learning up to $64\%$ faster than baselines in environments where standard approaches fail to leverage expert guidance effectively. 
Additionally, comprehensive parameter sensitivity analysis reveals the framework's robustness across varying dataset sizes, similarity thresholds, and confidence mechanism configurations.

This work demonstrates that accessible, suboptimal, observation-only expert data can substantially accelerate reinforcement learning when properly integrated. 
By removing the requirements for optimal expert performance and complete action information, our framework significantly expands the practical applicability of imitation learning to real-world scenarios where such data is expensive or impossible to obtain. 
The heterogeneous action space capability further enables knowledge transfer across agents with different physical capabilities, opening pathways for progressive robot learning and cross-platform policy transfer.\\

While DIIQN demonstrates strong performance across tested environments, several limitations warrant consideration. \\

The \textbf{restriction to discrete action spaces} limits applicability to many real-world domains. 
Extending DIIQN to continuous action spaces~\cite{lillicrap2015continuous,schulman2017proximal} would require adapting the action inference mechanism to handle continuous action distributions, potentially through density estimation techniques or by discretizing the continuous space adaptively.

The confidence mechanism currently relies solely on Q-value estimates and \textbf{training frequency by counter} to assess expert guidance quality. 
Other means of assessing the state-space coverage could be implemented.
Integrating intrinsic motivation techniques such as Random Network Distillation~\cite{burda2018exploration} (RND) could enhance the framework by providing additional signals about state-space coverage. 
The confidence mechanism could incorporate exploration bonuses to delay expert influence in novel regions until sufficient exploration has occurred, while maintaining aggressive expert utilization in well-understood areas. 
This would address the early-phase underutilization issue while maintaining the framework's cautious quality guarantees.

The bridging mechanism's computational demands stem from \textbf{exhaustive search through replay memory} for trajectory construction. 
Even though we have given helpful tips on how to alleviate this problem, it is clear that this brute-force approach becomes prohibitively expensive as memory size grows. 
A graph-based representation of the state space~\cite{replay_graph} could dramatically improve efficiency by pre-computing connectivity between frequently visited states. 
Nodes would represent state clusters, with edges denoting feasible single-step transitions discovered through agent experience. 
Bridge identification would then reduce to graph path-finding problems. 
Alternatively, learning a forward dynamics model could predict bridge existence without explicit search, though this introduces model error as an additional consideration.

The framework's distance metric dependency requires manual specification for each environment when the \textbf{state space cannot be normalized}, limiting generalization. 
Learning environment-specific distance metrics through metric learning approaches could automate this process~\cite{zhang2020learning}, potentially using contrastive learning on state transitions to discover meaningful similarity measures. 
This would reduce manual tuning requirements and improve transferability across related environments.


\appendix

\section{Distance Metric Selection}
\label{apdx:distances}
The choice of distance metric for state comparison depends on the characteristics of the environment's observation space. 
In this section, we detail the distance metrics employed for our experimental environments, focusing on Point Maze and the MinAtar suite, which represent different types of state representations requiring distinct similarity measures.

\paragraph{Point Maze} 
For Point Maze and other environments with low-dimensional state spaces, we utilize normalized state features as input to our model. 
Normalization is typically straightforward since minimum and maximum values are often known or can be empirically determined. 
In cases where outliers or extreme values occur, preprocessing through clipping to reasonable bounds or log transformation prior to normalization ensures stable metric behaviour.

Following normalization, each state dimension is bounded in $[0, 1]$, enabling consistent distance comparisons throughout the DIIQN procedure (action inference, KNN search, similarity filtering). 
We employ Euclidean distance as the metric for KNN search. 
Given the normalized state space with dimensionality $n$, the maximum possible distance between any two states occurs at opposite corners of the hypercube and is computed as:
\begin{equation}
  d_{max} = \sqrt{\sum_{i=1}^{n} (s_i^{max} - s_i^{min})^2} = \sqrt{\sum_{i=1}^{n} (1 - 0)^2} = \sqrt{n}  
\end{equation}

This maximum distance serves as the normalization constant for the similarity threshold $\tau_{similar}$, enabling threshold specification as a percentage (e.g., states within 90\% similarity correspond to distance $\leq 0.9 \cdot \sqrt{n}$). 
This percentile-based approach provides intuitive parameterization and ensures consistent behaviour across different state dimensionalities (see Section~\ref{sec:expert_sampling} for details on similarity filtering).

\paragraph{MinAtar} 
MinAtar environments require a specialized approach due to their multi-layered pixel-based state representation. 
Each state consists of multiple $10 \times 10$ channels, where each channel encodes a specific game object type (player, player projectiles, enemies, enemy projectiles, etc.). 
For normalization, pixel values can be scaled by dividing by $255$ for $8$-bit color representations, after which Euclidean distance can be applied. 
Alternatively, when pixels represent binary presence/absence without color information affecting gameplay, Hamming distance provides a computationally efficient metric.

However, a critical challenge arises in MinAtar's layered representation: semantically important information, such as the agent's position, may occupy only one or two pixels within a channel. 
Computing distance as the simple sum of per-channel Hamming distances causes this crucial information to be overwhelmed by channels with many active pixels but lower semantic importance. 

To address this, we employ \textbf{channel-based weighting} that increases the influence of sparse, information-rich channels while reducing the contribution of dense, less critical channels. 
We present two approaches:

\textbf{Static Weighting:} Channel weights can be manually tuned based on domain knowledge, assigning higher weights to critical channels (e.g., player position) and lower weights to background elements. 
These weights are computed once during dataset preprocessing, ensuring computational efficiency during training.

\textbf{Dynamic Sparsity-Based Weighting:} For a more general, tuning-free solution, we compute weights adaptively based on channel sparsity. 
When comparing two states $s$ and $s'$, for each channel $c$ we count the active pixels in both states, compute their density as the fraction of total pixels, and derive the averaged sparsity:
\begin{equation}
    \rho_c = 1 - \frac{1}{2}\left(\frac{\text{active\_pixels}_c(s)}{100} + \frac{\text{active\_pixels}_c(s')}{100}\right)
\end{equation}
The channel weight is then computed as:
\begin{equation}
    w_c = w_{\text{base}} + \lambda \cdot \rho_c
\end{equation}
where $w_{\text{base}}$ is a baseline weight and $\lambda$ is a scaling factor controlling the influence of sparsity. 
The final weighted Hamming distance is:
\begin{equation}
    D(s, s') = \sum_{c=1}^{C} w_c \cdot \text{Hamming}_c(s, s')
\end{equation}
where $C$ is the number of channels. This approach automatically prioritizes channels with sparse but semantically important information without requiring manual tuning for each environment.

\section{Hyperparameters}

We provide the complete hyperparameter configurations used in our experimental evaluations for the MinAtar environments (Table~\ref{tab:hyperparameters_minatar}) for DIIQN evaluation and Point Maze (Table~\ref{tab:hyperparameters_point}) for HA-DIIQN evaluation. Additionally, we describe the computational hardware infrastructure utilized for conducting all experiments (Table~\ref{tab:hardware}).

\renewcommand{\arraystretch}{1.3}
\begin{tabularx}{\textwidth} { 
   >{\centering\arraybackslash}m{1.5cm} 
   >{\raggedright\arraybackslash}m{5.5cm} 
   >{\centering\arraybackslash}m{2.5cm} 
   >{\centering\arraybackslash}m{2.5cm} }
    \caption{Hyperparameters used in final experiments for MinAtar environments (DIIQN)}\label{tab:hyperparameters_minatar}\\
    \hline
    Symbol & Description & Introduced in & Value \\
    \hline
    \hline
     $T_{\text{train}}$ & Training Period & DQN & $5$M  \\
    \hline
     $|\mathcal{D}_a|$ & Buffer Size & DQN & $200$K  \\
    \hline
     $B$ & Batch Size & DQN & $32$  \\
    \hline
     $\alpha$ & Learning Rate & DQN & $5\times 10^{-5}$  \\
    \hline
    $\epsilon$ & Exploration & DQN & $1 \to 0.01$ \newline ($100$K steps) \\
    \hline
    $T_{\text{warmup}}$ & Warmup Period & DIIQN & $20$K  \\
    \hline
    $\gamma$ & Discount Factor & DQN & $0.99$  \\
    \hline
    $f_{\text{learn}}$ & Learning Frequency & DQN & $1$  \\
    \hline
    $f_{\text{target}}$ & Target Model Update Frequency & DQN & $1$K  \\
    \hline
    $\alpha_{\text{PER}}$ & PER highlight & DQN & $0.6$  \\
    \hline
    $\beta_{\text{PER}}$ & PER control & DQN & $0.4 \to 1$ \newline ($400$K steps)  \\
    \hline
    $w_{base}$ & Dynamic Weighting Base Weight & DIIQN & $1$  \\
    \hline
    $\lambda$ & Dynamic Weighting Scaling Factor & DIIQN & $2$  \\
    \hline
    $\rho_{max}$ & Dynamic Weighting Max Sparsity & DIIQN & $1$  \\
    \hline
    $\tau_{similar}$ & Similarity Threshold & DIIQN & $0.99$  \\
    \hline
    $|\mathcal{D}_e|$ & Expert Dataset Size & DIIQN & $100$K  \\
    \hline
    $c_{max}$ & Expert Regions Maximum Threshold & DIIQN & $150$K  \\
    \hline
    -- & Optimizer & DQN & Adam  \\
    \hline
    -- & Activation Function & DQN & ReLU  \\
    \hline
    -- & Layers Size & DQN & Conv($16, 3\times3$)  \newline FC($1024 \to 128$)  \newline FC($128 \to output$) \\
    \hline
    $k_{knn}$ & Number of $K$-nearest neighbours & DIIQN & $5$  \\
    \hline
\end{tabularx}

\renewcommand{\arraystretch}{1.3}
\begin{tabularx}{\textwidth} { 
   >{\centering\arraybackslash}m{1.5cm} 
   >{\raggedright\arraybackslash}m{5.5cm} 
   >{\centering\arraybackslash}m{2.5cm} 
   >{\centering\arraybackslash}m{2.5cm} }
    \caption{Hyperparameters used in final experiments for Point Maze environments (HA-DIIQN)}\label{tab:hyperparameters_point}\\
    \hline
    Symbol & Description & Introduced in & Value \\
    \hline
    \hline
     $T_{\text{train}}$ & Training Period & DQN & $300$K  \\
    \hline
     $|\mathcal{D}_a|$ & Buffer Size & DQN & $50$K  \\
    \hline
     $B$ & Batch Size & DQN & $32$  \\
    \hline
     $\alpha$ & Learning Rate & DQN & $6.3\times 10^{-4}$  \\
    \hline
    $\epsilon$ & Exploration & DQN & $1 \to 0.05$ \newline ($100$K steps) \\
    \hline
    $T_{\text{warmup}}$ & Warmup Period & DIIQN & $5$K  \\
    \hline
    $\gamma$ & Discount Factor & DQN & $0.98$  \\
    \hline
    $f_{\text{learn}}$ & Learning Frequency & DQN & $1$  \\
    \hline
    $f_{\text{target}}$ & Target Model Update Frequency & DQN & $1$K  \\
    \hline
    $\alpha_{\text{PER}}$ & PER highlight & DQN & $0.6$  \\
    \hline
    $\beta_{\text{PER}}$ & PER control & DQN & $0.4 \to 1$ \newline ($200$K steps)  \\
    \hline
    $|\mathcal{D}_e|$ & Expert Dataset Size & DIIQN & $30$K  \\
    \hline
    $\tau_{similar}$ & Similarity Threshold & DIIQN & $0.96$  \\
    \hline
    $c_{max}$ & Expert Regions Maximum Threshold & DIIQN & $50$K  \\
    \hline
    -- & Optimizer & DQN & Adam  \\
    \hline
    -- & Activation Function & DQN & ReLU  \\
    \hline
    -- & Layers Size & DQN & FC($128 \to 64$)  \newline FC($64 \to output$) \\
    \hline
    $k_{knn}$ & Number of $K$-nearest neighbours & DIIQN & $5$  \\
    \hline
    $\tau_{infeas}$ & Infeasibility threshold & HA-DIIQN & $0.95$  \\
    \hline
    $k$ & Bridge Discovery Maximum Agent Depth & HA-DIIQN & $4$  \\
    \hline
    $n$ & Bridge Discovery Maximum Expert Depth & HA-DIIQN & $3$  \\
    \hline
\end{tabularx}

\renewcommand{\arraystretch}{1.3}
\begin{tabularx}{\textwidth} { 
   >{\raggedright\arraybackslash}m{5cm} 
   >{\raggedleft\arraybackslash}m{4.5cm} }
    \caption{Hardware Infrastructure}\label{tab:hardware}\\
    \hline
    Component & Specification \\
    \hline
    \hline
    Central Processing Unit (CPU) & Intel Core i7-9700 \\
    \hline
    Graphics Processing Unit (GPU) & GeForce RTX 2080 \\
    \hline
    Random-Access Memory (RAM) & $32$GB \\
    \hline
    Operating System & Ubuntu 22.04.4 LTS \\
    \hline
\end{tabularx}
\renewcommand{\arraystretch}{1}

\printcredits

\bibliographystyle{cas-model2-names}

\bibliography{sample-base}

\end{document}